\pdfoutput=1
\documentclass{article}
\usepackage{microtype}
\usepackage{graphicx}
\usepackage{subfigure}
\usepackage{booktabs} 

\usepackage{hyperref} 

\usepackage{xcolor}
\usepackage{amsmath}
\usepackage{amsfonts}
\usepackage{rotating}
\usepackage{multirow}
\newcommand{\cut}[1]{}
\newcommand{\nice}[1]{}
\newcommand{\doublecheck}[1]{\textcolor{black}{#1}}
\newcommand{\todo}[1]{\textcolor{red}{(#1)}}
\newcommand{\hblue}[1]{\textcolor{blue}{#1}}
\newcommand{\hgreen}[1]{\textcolor{green}{#1}}
\newcommand{\hidden}[1]{}
\newcommand{\keypoint}[1]{\textbf{#1}\quad}
\usepackage{array}
\newcolumntype{H}{>{\setbox0=\hbox\bgroup}c<{\egroup}@{}}
\newcolumntype{Z}{>{\setbox0=\hbox\bgroup}c<{\egroup}@{\hspace*{-\tabcolsep}}}

\DeclareMathOperator*{\argmax}{\arg\max}

\usepackage[accepted]{icml2019}

\icmltitlerunning{Investigating Generalisation in Continuous Deep Reinforcement Learning}

\begin{document}

\twocolumn[
\icmltitle{Investigating Generalisation in Continuous Deep Reinforcement Learning}




\begin{icmlauthorlist}
\icmlauthor{Chenyang Zhao}{ed}
\icmlauthor{Olivier Sigaud}{fr}
\icmlauthor{Freek Stulp}{de}
\icmlauthor{Timothy M. Hospedales}{ed}

\end{icmlauthorlist}

\icmlaffiliation{ed}{School of Informatics, University of Edinburgh, Edinburgh, United Kingdom}
\icmlaffiliation{fr}{Institut des Syst\`emes Intelligents et de Robotique, Sorbonne Universit\'e, Paris, France}
\icmlaffiliation{de}{Institute of Robotics and Mechatronics, German Aerospace Center (DLR), Wessling, German}
\icmlcorrespondingauthor{Chenyang Zhao}{c.zhao@ed.ac.uk}

\icmlkeywords{Machine Learning, ICML}

\vskip 0.3in
]



\printAffiliationsAndNotice{} 

\begin{abstract}
Deep Reinforcement Learning has shown great success in a variety of control tasks. However, it is unclear how close we are to the vision of putting Deep RL into practice to solve real world problems. In particular, common practice in the field is to train policies on largely deterministic simulators and to evaluate algorithms through training performance alone, without a train/test distinction to ensure models generalise and are not overfitted. Moreover, it is not standard practice to check for generalisation under \emph{domain shift}, although robustness to such system change between training and testing would be necessary for real-world Deep RL control, for example, in robotics. In this paper we study these issues by first characterising the sources of uncertainty that provide generalisation challenges in Deep RL. We then provide a new benchmark and thorough empirical evaluation of generalisation challenges for state of the art Deep RL methods. In particular, we show that, if generalisation is the goal, then common practice of evaluating algorithms based on their training performance leads to the wrong conclusions about algorithm choice. Finally, we evaluate several techniques for improving generalisation and draw conclusions about the most robust techniques to date.
\end{abstract}


\section{Introduction}
Deep Reinforcement Learning (Deep RL) has achieved great success in solving many complex problems ranging from discrete control tasks like Go \citep{silver2017mastering} and Atari games \citep{mnih2015human}, to continuous robot control tasks \citep{lillicrap2015continuous}. As intelligent systems, we would like our Deep RL agents to succeed in various environments, including ones unseen during training. However, as examples of high capacity machine learning models, Deep RL agents are at risk of \emph{overfitting} -- learning policies overly specific to their training environment and failing to generalise to new conditions. Overfitting risk, regularisation and generalisation to novel samples is well studied in supervised learning, where evaluating generalisation through train/test splits is ubiquitous. However (due to the relatively greater difficulty of obtaining a good solution to the RL training problem in the first place) evaluating for generalisation to novel conditions through such train/test splits is not common practice in Deep RL. Correspondingly,  mainstream Deep RL algorithm research focuses on optimising the training condition well, rather than developing models that generalise well to novel  conditions.  Nevertheless, now that Deep RL \emph{training} is increasingly successful, it is timely to move focus onto models' generalisation properties. Achieving generalisation is crucial if Deep RL should move out of the the lab and solve real-world problems where noise and uncertainty are intrinsic, and novel conditions will certainly be encountered \cite{sunderhauf2018limits}.  

In this paper, along with several concurrent works \cite{zhang2018dissection,packer2018assessing,zhang2018study}, we advocate for a renewed focus on generalisation in Deep RL research, rather than on training performance alone. Aside from first principles interest in the ability of our agents to succeed in diverse and novel environments, there is particular demand for generalisation in the context of the reality gap in robotics.  Despite continuing improvements in algorithmic sample efficiency, Deep RL requires a large number of environmental interaction samples for learning complex tasks without prior knowledge. For this reason, the majority of Deep RL training is done in simulation, which is moreover usually deterministic. Transfer from such simulated training environments to potential real-world deployment is known as crossing the reality gap in robotics, and is well known to be difficult \cite{Koos2013transferability}, thus providing an important motivation for studying generalisation. 

As we will see, there are several generalisation challenges that can arise for RL control agents. The first is generalisation from a deterministic training environment to a noisy and uncertain testing environment, for example in the form of real sensor and actuator noise. Secondly, assuming we correctly model environmental variability in our training simulation, there is the question of whether an agent learns to generalise to future conditions drawn from the same distribution, or overfits to its specific training experiences \cite{zhang2018study}. Finally, there is the subtle but important point that no matter the effort applied to modelling environmental conditions and variability in simulated training, it is generally impossible to predict and accurately model the environmental conditions and variability an agent might encounter in the real world \cite{Koos2013transferability}. Therefore an important way to think about model generalisation is not only robustness to overfitting per-se, but generalisation under some level of \emph{domain shift}. In supervised learning, domain shift refers to changes in the data distribution which we would like a predictive model to be robust to, for example the type of camera in visual object recognition \cite{csurka2017domainAdaptationBook}. The corresponding notion in RL is that we would like our policy's success to be invariant to nuisance changes in the environment \cite{cully2015robotAdapt}. These could span both noise, for example sensor, actuator, and environmental noise; and variability, for example camera type, initial state of an agent, or mass of an objects being  manipulated.

In this paper, we study generalisation in Deep RL for continuous control, with a particular focus on robustness to domain shift between training and testing. We first provide a thorough characterisation of the diverse sources of uncertainty and variability  that provide generalisation challenges. Secondly, we provide a thorough evaluation of several state of the art Deep RL methods on several OpenAI Gym benchmarks \cite{openaigym} in terms of their generalisation properties, particularly across domain shift, with regards to these different sources of variability. In doing so we attempt to answer several questions including: 
\begin{itemize}
\parskip0pt
\itemsep0pt
\item \emph{How do state of the art algorithms generalise under different sources of uncertainty and domain shift?} 
\item \emph{Does standard practice of picking algorithms and architectures based on training return lead to selecting models with good generalisation?}
\item \emph{Can robustness be improved via simple modifications to existing methods?}
\end{itemize}
Our analysis shows that existing RL methods are generally vulnerable to overfitting, showing poor generalisation to testing. This is particularly so in cases of domain shift, for example transferring from a deterministic to stochastic simulation; or where system parameters such as robot mass are varied between training and deployment. Correspondingly, the standard practice of picking algorithms and architectures based on training performance leads to the wrong choice in terms of generalisation performance. Therefore, if generalisation is of interest, then benchmarks such as ours that test generalisation are recommended instead. Finally, we thoroughly evaluate several existing techniques that might improve generalisation, and report the most robust combination as a starting point for future work.


\section{Related Work}

\keypoint{Standardised and Reproducible Evaluations} Deep RL research has benefitted tremendously from  efforts on standardised environment models and benchmarks \cite{openaigym, tassa2018deepmind}. Building on this, a variety of Deep RL algorithms for continuous control were implemented and compared based on \emph{training return} in \citet{duan2016benchmarking}. However, these results have high variance, leading to concerns about reproducibility of conclusions and dependence on specific choice of training seeds  \cite{henderson2017deep}. This in turn has led to statistical power analysis to determine the sufficient number of random seeds to allow reliable comparison among algorithms  \cite{colas2018many}.

\keypoint{Generalisation and Overfitting} 
Recent concurrent work to ours has noted the overfitting risk in this standard practice of evaluation by training return. \citet{zhang2018study} studied overfitting of Deep RL in discrete maze tasks. Testing environments are generated with the same maze configuration but different initial positions as training. Deep RL algorithms were shown to suffer from overfitting to training configurations and memorise training scenarios. Similarly 
\citet{zhang2018dissection} proposed to formalise overfitting in continuous control by splitting random seeds for training and testing environments, and then diagnosing overfitting through the generalisation error -- the difference between average return in training and testing environments. \citet{cobbe2018quantifying} studied regularisation for improving generalisation across procedurally generated arcade environments in discrete control. All these studies showed improved generalisation when trained with more random seeds. 

\keypoint{Domain Shift in RL}
Overfitting models fail to generalise from training to testing data although they are drawn from the same underlying distribution. Domain-shift challenges a model trained in one domain to perform in a target domain with different statistics, which typically leads to a significant drop in performance. This challenge is unavoidable if we wish to apply Deep RL-trained models in the real world, as the reality gap \cite{Koos2013transferability} of modelling errors and the unpredictability of the unconstrained real world mean that training in simulation will always mismatch reality. Recent concurrent work to ours has also studied some limited facets of domain-shift including adding noise to observations or initial states during testing compared to the training simulation \cite{zhang2018dissection}. Meanwhile \citet{packer2018assessing} studied performance under train-test domain shift by modifying environmental parameters such as robot mass and length  modified to generate new domains. In discrete control, generalisation under adversarially designed noise has also been studied \cite{huang2017adversarial}.

\keypoint{Improving Generalisation}
Various techniques can potentially improve generalisation of RL-trained policies in the face of overfitting and domain-shift. These include training under adversarially designed noise \cite{mandlekar2017adversarially} and specially designed network architectures \cite{srouji2018structured}. While entropy-regularisation \cite{haarnoja2017soft} is topical in RL for improving training performance, we also investigate its potential for improving RL generalisation as it does in supervised learning \cite{chaudhar2017entropySGD}. Meanwhile domain randomised  training has been used to improve generalisation to new domains \cite{tobin2017domainRandomization,stulp2011uncertainGrasp,colas2018many}. 

\keypoint{Contributions} We present a more thorough characterisation of the generalisation challenges in overfitting and domain shift in continuous control compared to concurrent studies \cite{zhang2018study,packer2018assessing,zhang2018dissection}, which each only touch on a subset of the factors involved (Table~\ref{table:comparison}). To quantify these issues empirically, we contribute a comprehensive benchmark for measuring RL generalisation performance, both within and across-domain, and evaluate several current algorithms and modifications.


\section{Characterising Generalisation Challenges}

We first start by giving a thorough characterisation of the generalisation challenges that can arise for RL agents.

\keypoint{RL formalisation}
In reinforcement learning, an agent learns to maximise its expected cumulative reward through interacting with an environment. The problem setting of RL can be described by a Markov Decision Process $\mathcal{M = (S,A,P,} r ,\gamma,\mathcal{P}_0 )$, with state space $\mathcal{S}$, action space $\mathcal{A}$, environment (transition probability) model $\mathcal{P}(s_{t+1} | s_t, a_t) $, reward function $r(s_t, a_t, s_{t+1})$ and discount factor $\gamma \in [0,1]$. In addition, $\mathcal{P}_0$ describes the probability distribution of initial state $s_0 \in \mathcal{S}$. The objective of RL learning is to find the optimal policy $\pi^*: \mathcal{S} \rightarrow\mathcal{A}$ to act in an MDP such that,  
\begin{align}
\pi^* &= \argmax_{\pi} \mathbb{E}_{\tau \sim (\pi, \mathcal{P}, \mathcal{P}_0)} \left[  \sum_{t=0}^\infty \gamma^t r(s_t, a_t(\pi)) \right] \\
& = \argmax_{\pi} \mathbb{E}_{\tau \sim (\pi, \mathcal{P}, \mathcal{P}_0)}  \left[ R(\tau) \right]\label{eq:pointRet}
\end{align}
where trajectory $\tau$ is a sequence of state and action pairs $\{(s_0, a_0),...\}$, $R(\tau)$ is the accumulated return of trajectory $\tau$, and $\mathbb{E}_{\tau \sim (\pi, \mathcal{P}, \mathcal{P}_0)} \left[ \cdot \right]$ indicates that trajectory $\tau$ is sampled with $s_0 \sim P_0$, $a_t \sim \pi(\cdot | s_t)$, $s_{t+1} \sim \mathcal{P}(\cdot | s_t, a_t )$.

\subsection{Sources of Uncertainty and Variability}

\begin{figure}[t]
\centering
\includegraphics[width=0.8 \columnwidth]{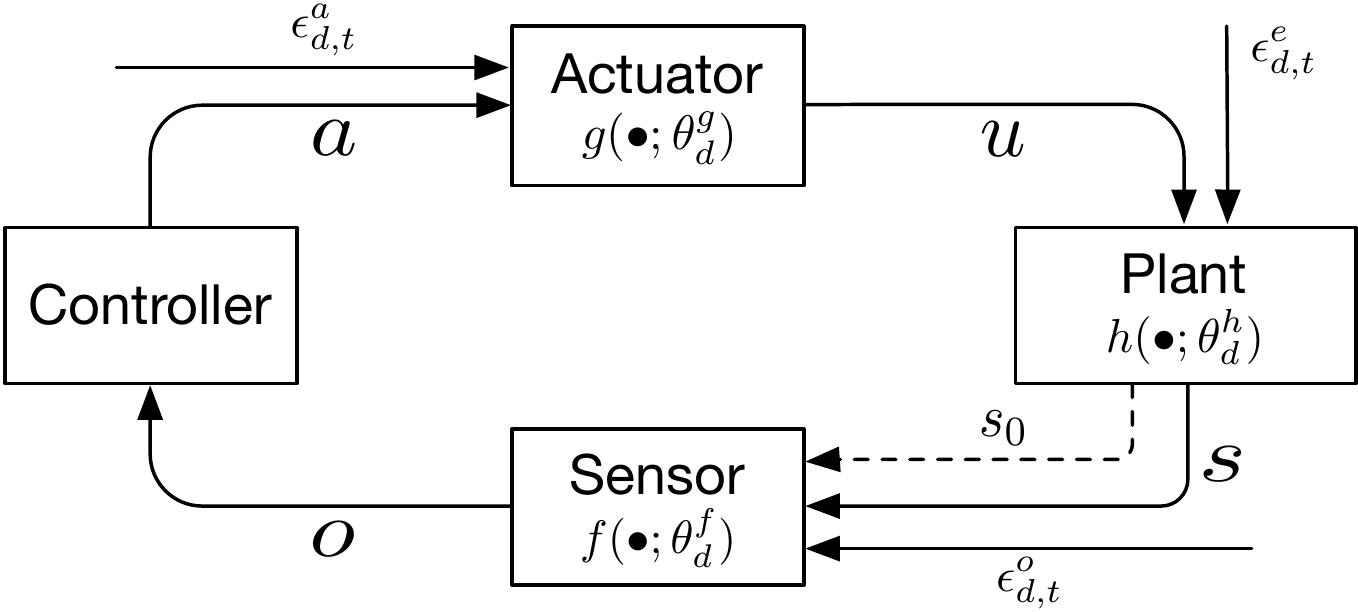}
\caption{Block diagram of a classic control system. $f$ represents the observation function, $g$ represents the actuation function, $h$ represents the \hidden{dynamic} transition function, and $\epsilon$ represents the noise that enters the system due to different types of uncertainties.} 
\label{fig:block}
\end{figure}

Consider a classic robot control system as illustrated in Figure~\ref{fig:block}.
The agent $\pi$ provides the controller and the environmental transition model can be broken into an actuation module, a sensor module and a dynamical module. In real world applications, each of the three modules may contain uncertainties due to noise. Moreover, each may exhibit contextual variability which provides a potential source of bias between distinct encounters with the environment, including between training and testing. Such variability creates \emph{domain shift} in RL. To unpack the distinction between uncertainty and disturbances in rollouts (i.e. noise), and contextual  variations that can induce systematic shifts between domains $d$, we specify the environmental MDP corresponding to Figure~\ref{fig:block} in more detail as Eq.~\eqref{eq:mdp1},\eqref{eq:mdp2}. Note that though the uncertainties are described as Gaussian distributions for simplicity, they can be any distribution in practice. 
\cut{\paragraph{Suggestions Freek}}
\begin{eqnarray}
p_d(s_0) \hspace{0cm}s_{d,0} & \quad \quad \sim \mathcal{N}(\mu^{s_0} + \theta_d^{s_0}, \sigma^{s_0}_d) \label{eq:mdp1}\\
p_d(s_{t+1}|s_{t},a_t)&\begin{cases}
o_{d,t} &\sim \mathcal{N}(f(s_{d,t}; \theta^f_d), \sigma_{d}^o)\\
u_{d,t} &\sim \mathcal{N}(g(a_{d,t}; \theta^g_d), \sigma_{d}^u)\\
s_{d,t+1} &\sim \mathcal{N}(h(s_{d,t},u_{d,t};\theta^h_d), \sigma_{d}^s)
 \end{cases} \label{eq:mdp2}
\end{eqnarray}
\hidden{Remark: $\sigma^u$ or $\sigma^a$?}

In this model of the MDP, noise is introduced at three time-scales.
The observations $o$, commands $u$ and next states $s$ as perturbed at {\it each time step}, as indicated by the ${}_t$ subscript. They are perturbed by Gaussian noise with variances $\sigma^o$, $\sigma^u$, and $\sigma^s$, respectively. 
The initial state $s_o$ is sampled {\it once per episode} from a Gaussian with variance $\sigma^{s_0}$,  before the episode starts.
The combined function parameters \doublecheck{$\theta^{s_0,f,g,h}$} and the corresponding variances \doublecheck{$\sigma^{s_0,o,u,s}$} define the MDP. Switching between these parameters implies a domain switch. As discussed in Section~\ref{section:3.2}, these can also be sampled from a distribution {\it before learning}. These parameters then stay fixed during learning, which is why they are indexed with domain $_d$. 

\keypoint{Examples} To illustrate these terms by way of example: A change in the mass of an object to be manipulated (or of the robot itself) in the environment or friction constant would correspond to a change in transition function parameter $\theta^h_d$. Stochasticity in the outcomes of transition model due to environmental noise such as changing wind condition is determined by $\sigma^s_{d}$.  Changes in the observation function $f$ via parameters $\theta^f_d$ correspond to events such as a change of camera when doing vision-driven control. Meanwhile proprioceptive noise is generated with variance $\sigma^o_{d}$. Changes in the actuation function $g$'s parameter $\theta_d^g$ could correspond to wear in a motor or increased joint friction reducing the obtained forces. Noise generated internally by motor while actuating actions are sampled with variance $\sigma^u_{d}$. In general, we would like our agents to be robust to as much of these variations and noise as possible. \nice{Include some citations to examples of these}

\cut{
\paragraph{ICML 2019 Version}

\begin{eqnarray}
p_d(s_0)& \hspace{0cm}s_{d,0} \hspace{0.7cm} = \mu^{s_0} + \theta_d^{s_0} \hspace{1.0cm} + \epsilon^{s_0}_d\\
p_d(s_{t+1}|s_{t},a_t)&\begin{cases}
o_{d,t} &= f(s_{d,t}; \theta^f_d) \hfill + \epsilon_{d,t}^o\\
u_{d,t} & = g(a_{d,t}; \theta^g_d) \hfill + \epsilon_{d,t}^a\\
 s_{d,t+1} &= h(s_{d,t},u_{d,t};\theta^h_d) \hfill + \epsilon_{d,t}^e
 \end{cases}
\end{eqnarray}

Here each update is determined by a term dependent on the context (domain) $d$ and a time-dependent noise $\epsilon$. $\theta_d$ are the environmental parameters that describe an MDP $d$, these are fixed within domains $d$ but potentially biased across domains. Note that while we write the noise processes as additive for illustrative convenience, this is not necessary for our argument. The noise samples are drawn at each timestep, e.g., in the case of gaussian noise, $\epsilon^o_{d,t} \sim \mathcal{N} (0, \sigma^o_{d})$ where variance $\sigma^o_{d}$ is another  MDP parameter. For simplicity, we summarise the domain-dependent parameters as $\Theta = \{ \langle \theta^f, \theta^g, \theta^h \rangle, \langle \sigma^o, \sigma^a, \sigma^e \rangle\}$. While we consider MDP paremeters $\Theta$ to be fixed here, we consider distributions over these parameters in the next section.

\keypoint{Examples} To illustrate these terms by way of example: A change in the mass of an object to be manipulated (or of the robot itself) in the environment or  friction constant would correspond to a change in transition function parameter $\theta^h_d$. Stochasticity in the outcomes of actions due to environmental noise is determined by samples $\epsilon^e_{d,t}$, and if one environment has more uncertain outcomes than another, this is determined by parameter $\sigma^e_d$.  Changes in the observation function $f$ via parameters $\theta^f_d$ correspond to events like a change of camera if doing vision-driven control. Meanwhile proprioceptive noise is modeled by samples $\epsilon^o_{d,t}$. Changes in the actuation function $g$'s parameter $\theta_d^g$ could correspond to wear in a motor or increased joint friction reducing the obtained forces.  In general, we would like our agents to be robust to as much of these variations and noise as possible. \nice{Include some citations to examples of these}

}
\begin{table}[bt]
\centering
\caption{Comparison of evaluation settings in: Basic Gym \cite{openaigym} Overfitting \cite{zhang2018study}, Dissection \cite{zhang2018dissection}, Generalize \cite{packer2018assessing}. $\{\cdot\}_{tr}$ are  samples generated with random training  seeds, $\{\cdot\}_{te}$ are  samples generated with random testing  seeds. }\label{table:comparison}
\vspace{-0.2cm}
\begin{center}
\begin{small}
\begin{sc}
\resizebox{1.0\columnwidth}{!}{
\begin{tabular}{c c c}
\toprule
Study & Train Setting & Test Setting \\
\midrule
Basic Gym  & $s_{0,tr} \sim \mathcal{N}(\cdot, \sigma)$  &  $s_{0,te} = s_{0, tr}$ \\
\midrule
Overfit. & $s_{0,tr} \sim \mathcal{N}(\cdot, \sigma)$  &  $s_{0,te} \sim \mathcal{N}(\cdot, \sigma)$ \\
\midrule
\multirow{4}{*}{Dissect.}& \multirow{2}{*}{$s_{0,tr} \sim \mathcal{N}(\cdot, \sigma_1)$} & $s_{0,te} \sim \mathcal{N}(\cdot, \sigma_1)\cut{\epsilon_{te}^{s_0} \neq \epsilon_{tr}^{s_0}}$\\
& 
&  $s_{0,te} \sim \mathcal{N}(\cdot, \sigma_2)$ \\
\\[-0.9em]
\cline{2-3}
\\[-0.9em]
& 
$s_{0,tr} \sim \mathcal{N}(\cdot, \sigma_1)$, &  $s_{0,te} \sim \mathcal{N}(\cdot, \sigma_1)\cut{\epsilon_{te}^{s_0} \neq \epsilon_{tr}^{s_0}}$, \\
& $o_{tr} \sim \mathcal{N}(\cdot, 0)$ \cut{$o_{tr} = \cdot$} & $o_{te} \sim \mathcal{N}(\cdot, \sigma_2)$ \\
\midrule
\multirow{4}{*}{Generaliz.} & \cut{\multirow{4}{*}{$s_{0,tr} \sim \mathcal{N}(\cdot, \sigma)$}} & $s_{0,te} \sim \mathcal{N}(\cdot, \sigma),\cut{,  \epsilon_{te}^{s_0} \neq \epsilon_{tr}^{s_0}}$\\
& $s_{0,tr} \sim \mathcal{N}(\cdot, \sigma)$, & $\theta_{te}^h \sim \mathcal{U}(\theta_0-\sigma_1, \theta_0+\sigma_1)$\\
& $\theta^h_{tr} = \theta_0$ & $s_{0,te} \sim \mathcal{N}(\cdot, \sigma)$,\\
& & $\theta_{te}^h \sim \mathcal{U}(\theta_0-\sigma_2, \theta_0+\sigma_2)$\\
\bottomrule
\end{tabular}}
\end{sc}
\end{small}
\end{center}
\vskip -0.1in
\end{table}

\subsection{Generalisation Across MDP Distributions}
\label{section:3.2}
With this formalisation in mind, we can understand the goal of generalisation as robustness to a potential \emph{distribution} of both environments $p(\Theta)$ and samples from those environments. That is, insensitivity to both environmental parameters $\Theta$, and noise samples. This is in contrast to commonly used deterministic simulations ($\sigma=0$), without environmental variability ($\Theta$ constant). 

\keypoint{Formalising Generalisation}
Given a fixed set of environmental parameters $\Theta$, the corresponding transition model and initial state distribution are denoted $\mathcal{P}^{\Theta}$ and $\mathcal{P}_0^{\Theta}$. We denote $\eta_{\Theta}(\pi) $ as the expected return given a set of environmental parameters $\Theta$. If the environmental parameters are varying across trials, we denote $\eta_{p(\Theta)}(\pi)$ as the expected return under the distribution of environment parameters $p(\Theta)$:
\begin{align}
\eta_{\Theta}(\pi) &= \mathbb{E}_{\tau \sim (\pi, \mathcal{P}^{\Theta}, \mathcal{P}_0^{\Theta})} \left[ R(\tau) \right] \label{eq:domainRet} \\
\eta_{p(\Theta)}(\pi) &=  \mathbb{E}_{\Theta \sim p(\Theta)} \left[ \eta_{\Theta}(\pi) \right].\label{eq:distRet}
\end{align}
We would like our agents to solve a distribution over (non-deterministic \doublecheck{$\sigma^{(o,u,s)}>0$)} environments in Eq.~\eqref{eq:distRet}, rather than the conventional RL criterion in Eq.~\eqref{eq:pointRet}. {That is for a given trial, we would expect to sample an environment once $\Theta\sim\mathcal{N}(\Theta_0,\Sigma)$, and then at each time-step sample noise $\epsilon\sim\mathcal{N}(0,\sigma)$. We want agents to perform well over both this long-timescale variability, and short-time scale uncertainty.}

Furthermore, by evaluating on training return, standard RL practice implicitly assumes that the simulated training domain models the testing domain perfectly: $\Theta_{tr} = \Theta_{te}$ or $p_{tr}(\Theta)=p_{te}(\Theta)$. While this assumption can hold for some tasks like Atari games, creating a sufficiently accurate simulated model is challenging for dynamic tasks \cite{Koos2013transferability}, and is generally impossible if the testing domain is the unconstrained real-world. Therefore an important quantity of interest to measure is how trained models generalize to encounters with a certain degree of domain shift between environments ($\Theta_{tr} \neq \Theta_{te}$ or $p_{tr}(\Theta)\neq p_{te}(\Theta)$). Therefore, besides training performance, we should monitor the robustness of our models via the quantity:
\begin{align}
\eta_{p_{te}(\Theta)}(\pi^*) ~|~ \{\pi^* =\argmax_{\pi} \eta_{p_{tr}(\Theta)}(\pi),~p_{te}(\Theta)\neq p_{tr}(\Theta)\}. \label{eq:distShift}
\end{align}
That is, the performance of the model $\pi^*$ trained on $\Theta_{tr}$ or $p_{tr}(\Theta)$; when tested on $\Theta_{te}$ or $p_{te}(\Theta)$. This view encompasses robustness to changes in distribution of starting condition \cite{zhang2018study} and maps \cite{cobbe2018quantifying}, training with deterministic observations $\to$ non-deterministic testing \cite{zhang2018dissection} (but also includes action and environmental noise), and extrapolation in environmental parameters such as object mass \cite{packer2018assessing} that will arise in the practice due to the reality gap \cite{Koos2013transferability}. Given the inability to exactly control or simulate the distribution of real-world environmental encounters, the model robustness quantified above should be a consideration in our development of new methods, and our choice of algorithms and architectures in practice. 

Since the vast majority of existing work does not explicitly separate training and testing phases, in the following sections we introduce a set of benchmarks with clear train/test distinctions.  Based on these, we systematically measure the generalization of several popular algorithms under the diverse variations discussed in this section.


\section{Experimental Design}

We design a benchmark of generalisation -- testing rather than training performance. We cover  both generalisation across seeds when the simulation is non-deterministic ($\sigma^{(o,u,s)}>0$   ) in observation, actuation and process; and particularly focus on robustness to environment parameter variation, i.e., domain-shift $\Theta_{tr}\neq\Theta_{te}$ or $p_{tr}(\Theta)\neq p_{te}(\Theta)$. 

\subsection{Training Algorithms and Architectures}\label{sec:algorithms}

We study several model-free policy gradient based Deep RL algorithms with OpenAI baseline implementations \cite{baselines} including Trust Region Policy Optimisation (TRPO) \cite{schulman2015trust}, Proximal Policy Optimisation (PPO) \cite{schulman2017proximal} and Deep Deterministic Policy Gradient (DDPG) \cite{lillicrap2015continuous}. 
In addition to basic Deep RL algorithms, we also consider several modifications of the baseline algorithms and architectures that may improve generalisation of learned policies. 

\keypoint{Entropy Regulariser}
Policies with higher entropy may be more robust to uncertain dynamics \cite{ziebart2010modeling}. To encourage learning higher-entropy policies, we add entropy to the training objective as a regularizer \cite{haarnoja2017soft}, denoted with suffix \textit{-Ent}. \cut{In entropy regularised learning \textit{PPO-ent} entropy regulariser coefficient is $0.001$.} 

\keypoint{Structured Control Net}
Inspired by classic control theory, Structured Control Net (SCN) splits a Deep RL policy into a linear module and a nonlinear residual module and shows improved robustness against noise \cite{srouji2018structured}. We train SCNs with PPO, denoted \textit{PPO-SCN}. 

\keypoint{Architecture}
A standard continuous control policy architecture is a multilayer perceptrons (MLP) with two 64 unit hidden layers\cut{ (except for SCN, which uses another linear layer to fuse the modules)}. To investigate the influence of network size on generalisation performance, we use a smaller MLP with two 16 unit hidden layers. {The  smaller network is indicated as with suffix \textit{-16}, such as \textit{PPO-16}, \textit{SCN-16}. }

\keypoint{Adversarial Attacks Assist Learning} 
Several attempts have been made to improve policy robustness with the assistance of adversaries \cite{pattanaik2018robust, pinto2017robust}. We follow `adversarially robust policy learning' (ARPL) \cite{mandlekar2017adversarially} where adversarial  noise maximises the norm of output actions $\delta_t = \epsilon \nabla_s || \pi_{\theta}(s_t) ||
$. 
\cut{as defined in Eq.~\eqref{eq: adv}.
\begin{equation}
\delta_t = \epsilon \nabla_s || \pi_{\theta}(s_t) ||
\label{eq: adv}
\end{equation}}
\cut{In the case of stochastic policy, we use the output distribution mean.} To minimise the interference in the simulation platform, adversaries only attack in the observation space. 

\keypoint{Multi-Domain Learning} 
Training agents on multiple domains is a simple strategy to improve generalisation over environment changes \cite{tobin2017domainRandomization}. {To simulate variability in domains, we generate a distribution of domains controlled by a parameter $\Sigma$. At each training rollout,  we sample a new domain from the distribution $\Theta\sim~\mathcal{N}(\Theta_0,\Sigma)$}. In this case we only sample dynamics parameters $\theta^h\in\Theta$, and denote this training setting \textit{-MDL}.

\subsection{Environments and Evaluation}
We experiment on several MuJoCo simulated \cite{todorov2012mujoco} environments in OpenAI Gym \cite{openaigym}. To explore robustness to environmental parameter variation, we modify various environmental dynamics parameters $\theta^h$ as shown in Table~\ref{table:env}.  For example, in \textit{Walker2d}, we modify robot mass, friction and gravity coefficients, and apply constant horizontal force as wind.

\begin{table}[tb]
\centering
\caption{Summary of evaluation environments and their environment factors that are included to generate shifts in transition model.}
\label{table:env}
\begin{center}
\begin{small}
\resizebox{1.0\columnwidth}{!}{
\begin{tabular}{l l}
\toprule 
Task Name & Environment Factors\\
\midrule
InvertedPendulum-v2 & $m_{cart}$, $m_{pole}$\\
InvertedDoublePendulum-v2 & $m_{cart}$, $m_{pole_1}$, $m_{pole_2}$\\
Walker2d-v2 & $m_{body}$, $c_{wind}$, $c_{friction}$, $g$ \\
Hopper-v2 & $m_{body}$, $c_{wind}$, $c_{friction}$, $g$ \\
HalfCheetah-v2 & $m_{body}$, $c_{friction}$, $g$ \\
\bottomrule
\end{tabular}}
\end{small}
\end{center}
\end{table}

For each training setting (task, algorithm/architecture, train-environment), we train 12 policies with different random seeds. For each condition (task, algorithm/architecture, test-environment) we evaluate by averaging over 20 different testing rollouts.
We assume constant sensor and actuation module context parameters  $\theta_{tr}^{(f,g)} = \theta_{te}^{(f,g)}$, and the same initial state distribution but different random seeds between training and testing $p_{tr}(s_0)~=~p_{te}(s_0)\cut{, \epsilon_{te}^{s_0} \neq \epsilon_{tr}^{s_0}}$. Our evaluation focuses on robustness against various noise scales in observation, action and environmental parameter space $\sigma_d^o, \sigma_d^u, \sigma_d^s$, as well as systematic shifts in the environmental parameters $\theta_d^h$. Gaussian noise is directly added to outputs of dynamic plant and agent policies as observation and action noise. For environmental parameter noise, at each time step, a set of environmental parameter (e.g. wind condition) is sampled and simulation is modified accordingly. In contrast, for systematic shifts, environmental parameters are sampled at the start of each trial and remain constants within the trial. Four testing settings are denoted with \textit{Obs, Act, Env, Dom} respectively.

\keypoint{Metrics} We use three evaluation metrics including \emph{Testing return} (Eq.~\eqref{eq:domainRet}) $\eta_{\Theta_{te}}(\pi)$ where possibly $\Theta_{te}\neq\Theta_{tr}$ and \emph{Expected testing return} $\eta_{p_{te}(\Theta)}(\pi)$ (Eq.~\eqref{eq:distRet}) where possibly  $p_{te}(\Theta)\neq p_{tr}(\Theta)$. Finally, as an aggregate measure of performance given that we may not know the strength of noise or variability in the testing domain, we also compute the \emph{Area Under Curve (AUC)} of testing return with respect to scale of the underlying Gaussian distribution $[\sigma_1, \sigma_2, ..., \sigma_N]$:
\begin{equation}
\text{AUC}(\pi) = \sum_{n=1}^N \Delta \sigma \eta_{\sigma_n}(\pi). \label{eq:AUC}
\end{equation}
{where $\sigma$ could be both noise \doublecheck{$\sigma^{(o,u,s)}$} and domain shift $\Sigma$, and $\Delta \sigma$ is the step size of varying scales, $\Delta \sigma = \sigma_2 - \sigma_1$.}


\section{Experimental Results}

\keypoint{How do standard continuous controllers generalise under noise and domain-shift?}

\begin{figure*}[t]
\centering
\subfigure[Observation Noise $\sigma^o$]
{\includegraphics[width=.49\columnwidth]{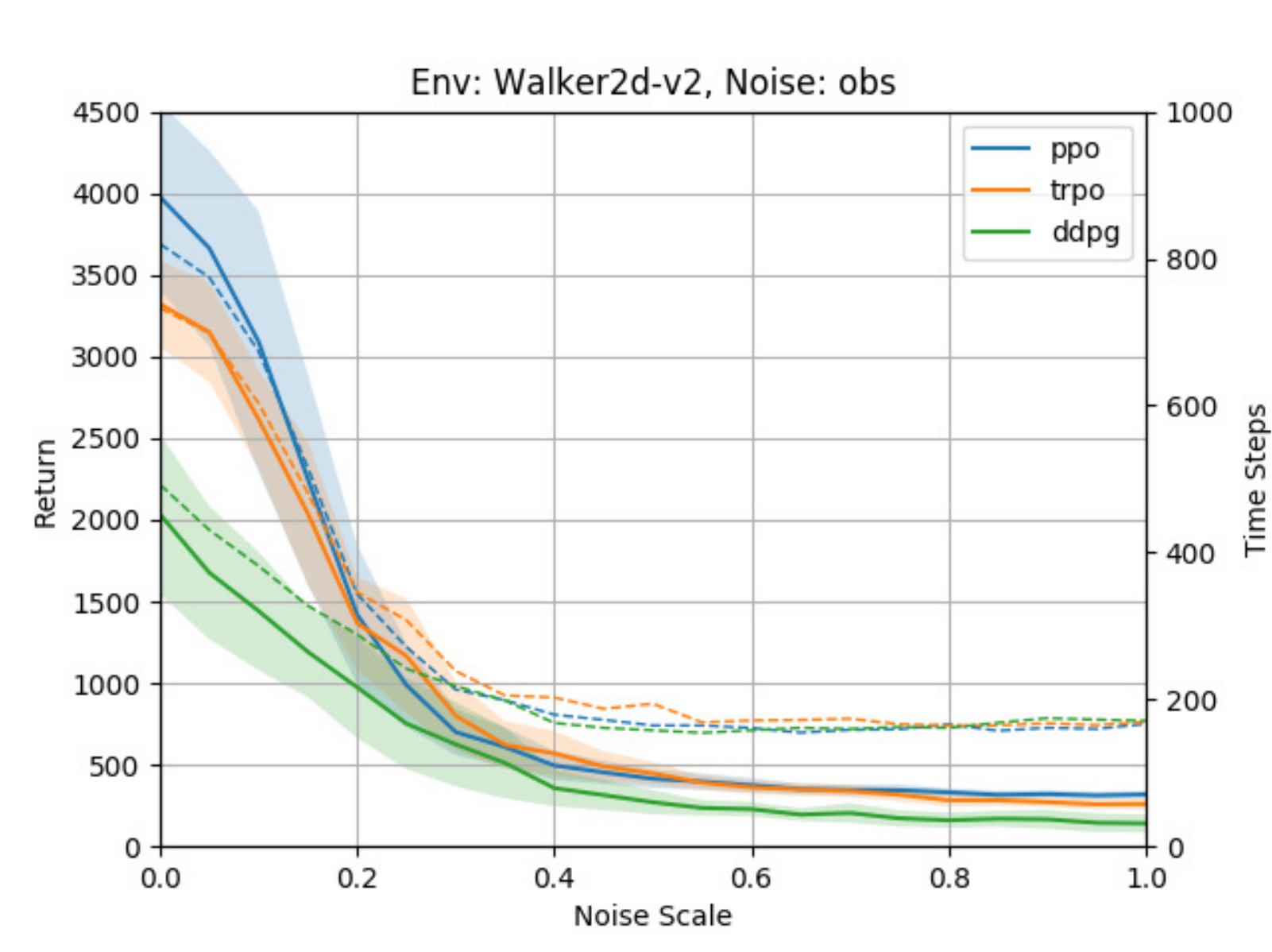}}
\subfigure[Action Noise  $\sigma^u$]
{\includegraphics[width=.49\columnwidth]{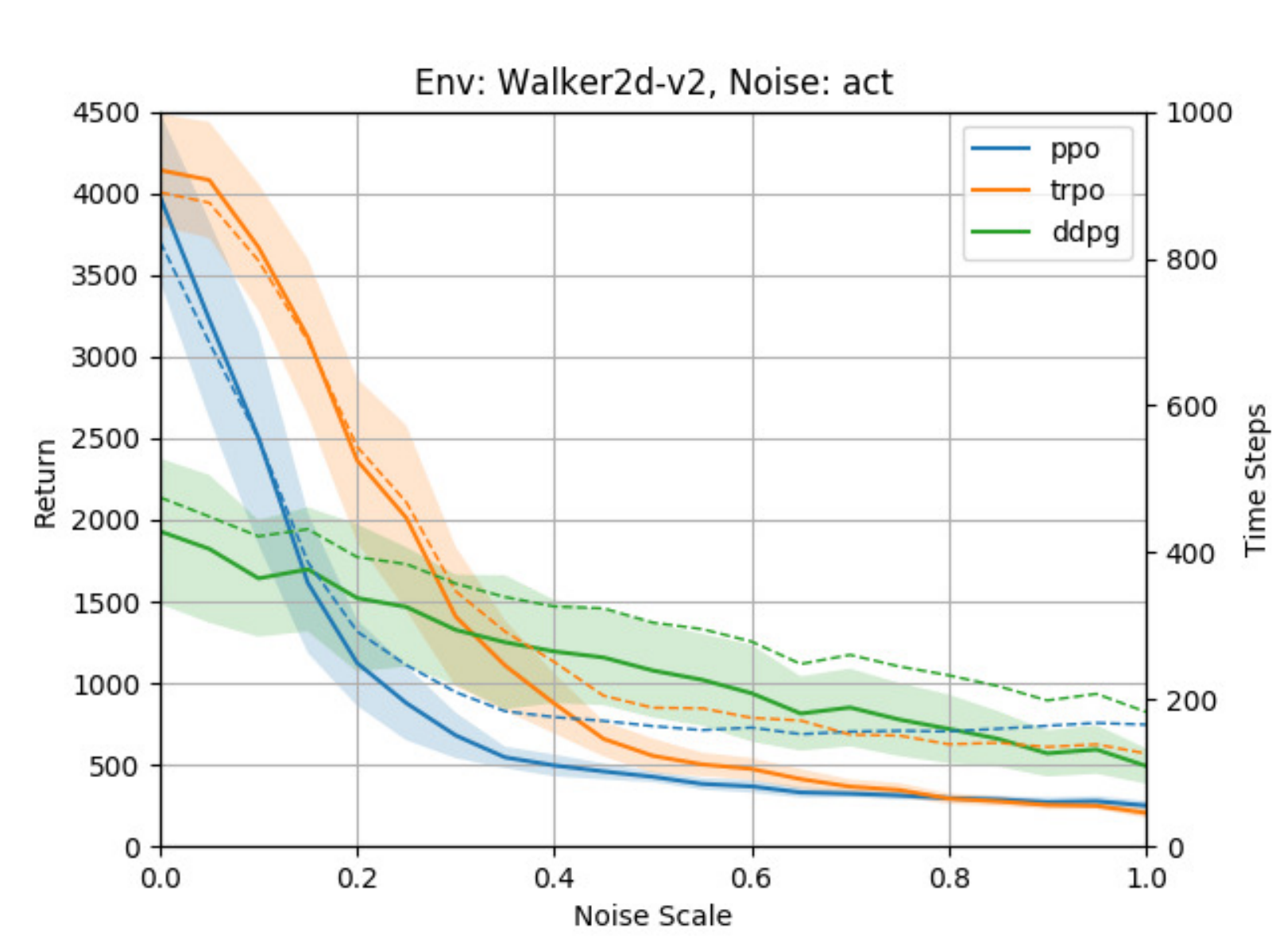}}
\subfigure[Environmental Noise  $\sigma^s$]
{\includegraphics[width=.49\columnwidth]{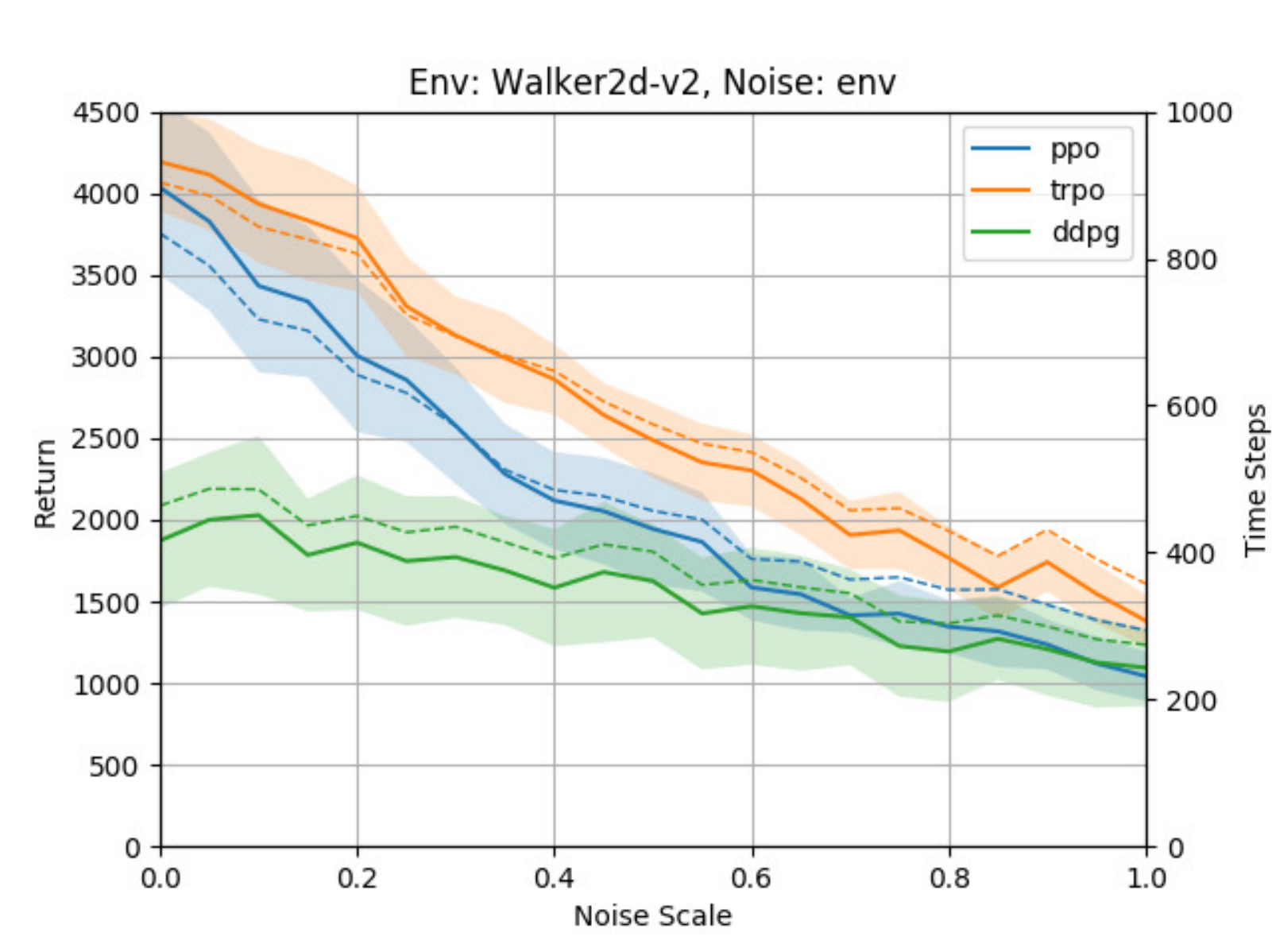}}
\subfigure[Domain Shift  $\Sigma$]
{\includegraphics[width=.49\columnwidth]{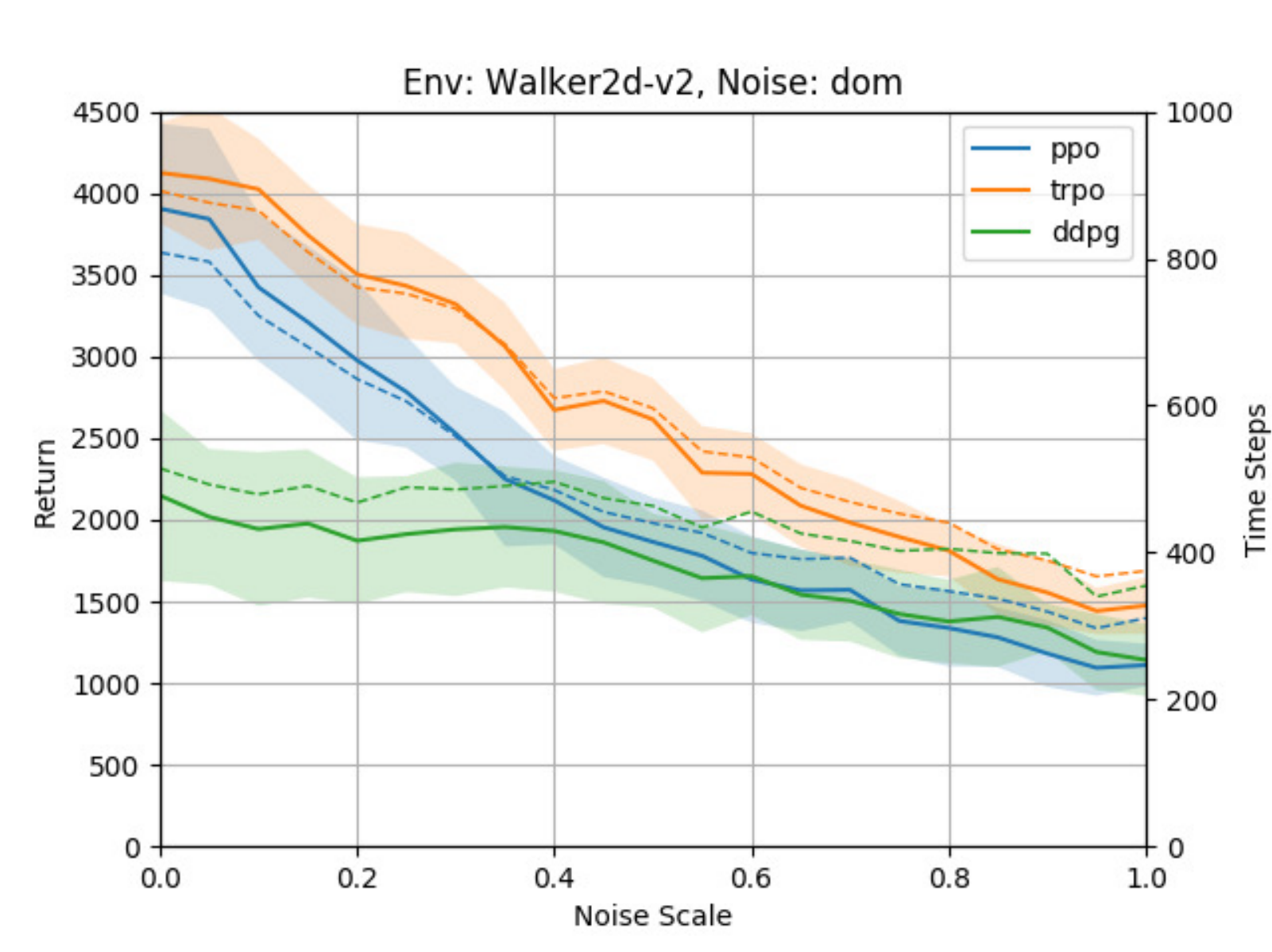}}
\vskip -0.1in
\centering
\subfigure[PPO]
{\includegraphics[width=.55\columnwidth]{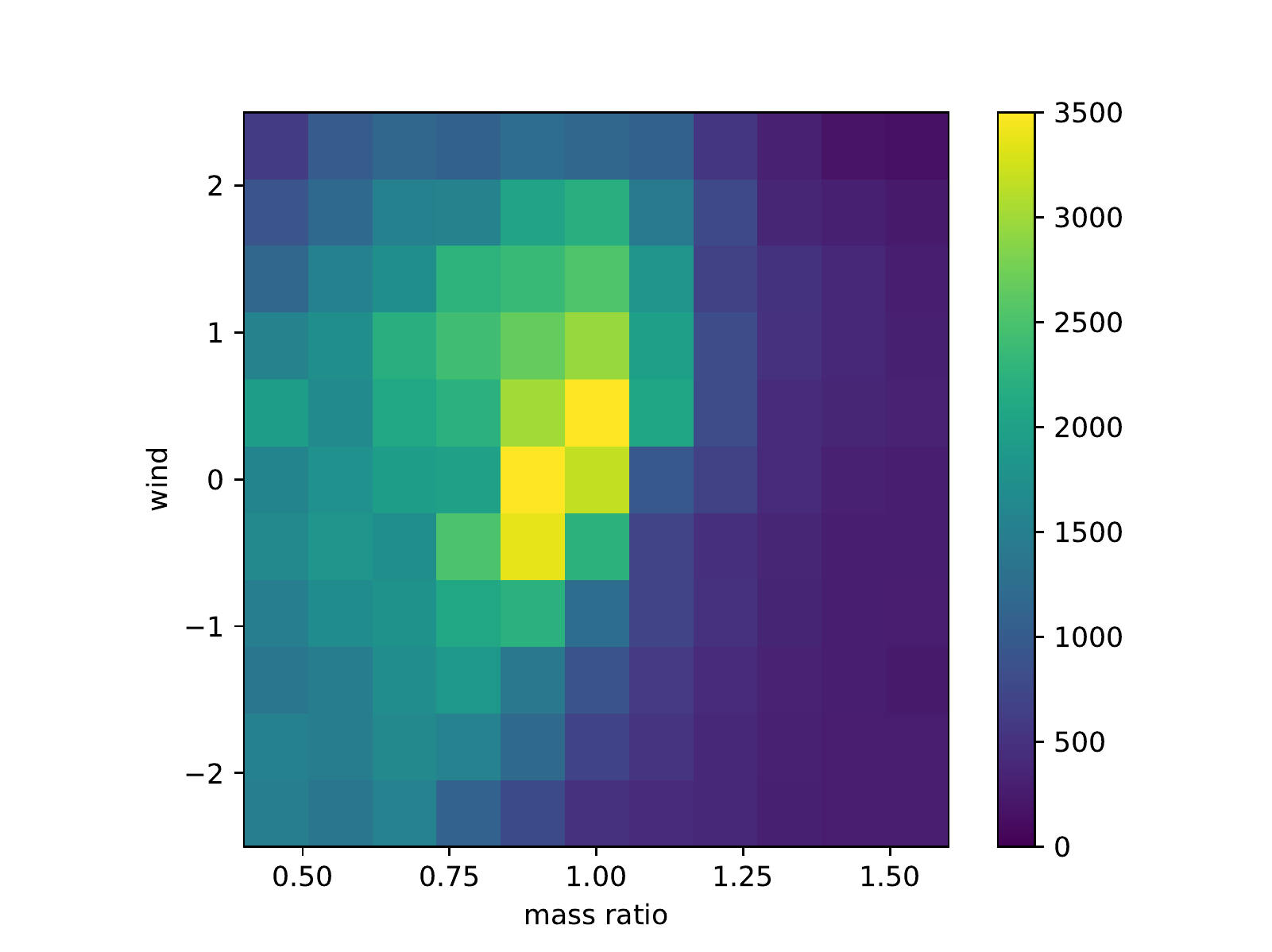}}
\subfigure[TRPO]
{\includegraphics[width=.55\columnwidth]{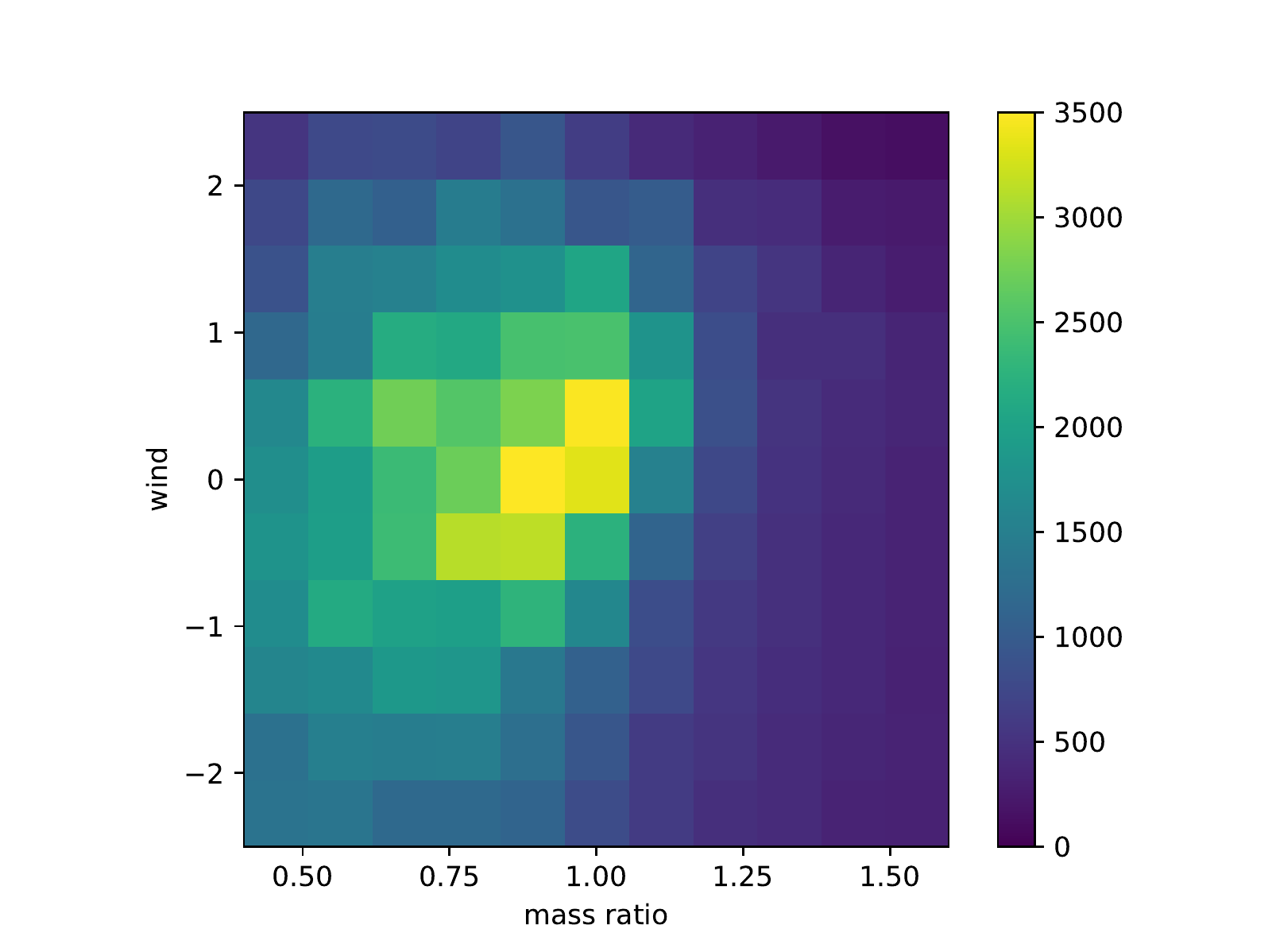}}
\subfigure[DDPG]
{\includegraphics[width=.55\columnwidth]{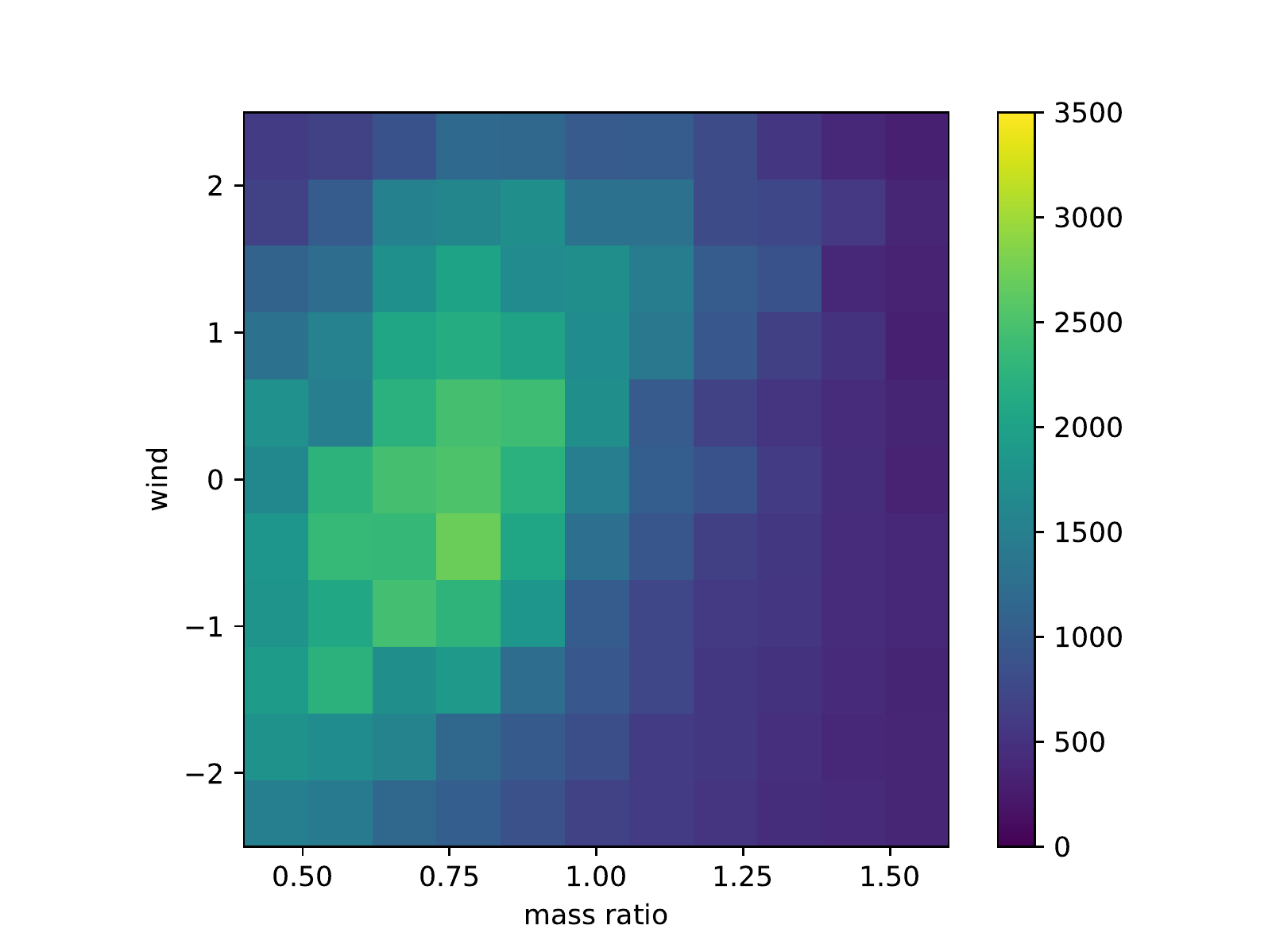}}
\vskip -0.1in
\caption{Generalization of standard continuous control policies for \textit{Walker2d-v2}. Top: Performance with varying testing noise $\sigma^{(o,u,s)}$ and  environment variation $\Sigma$ scale. Bottom: Heatmaps illustrate policy performance over a grid of environmental domain-shifts. Each cell corresponds to a particular set of context parameters $\theta_{te}^h$ with training domain at $(0,0)$. Results are averaged over 12 random seeds.}
\label{fig: ch5/obsact}
\end{figure*}

Standard practice in Deep RL research is to train policies on standard  MuJoCo-Gym benchmark environments. Algorithms such as TRPO and PPO now achieve impressive training return; but do the resulting policies generalise to novel contexts at testing time? 

We analyse this question for observation-, action-, and environment-noise and domain shifts. Figure~\ref{fig: ch5/obsact} shows the results for \textit{Walker2d} as an example environment, and the full results are summarised in Table~\ref{table: network1}\cut{(columns:  PPO, TRPO, DDPG)} in Appendix. Figures~\ref{fig: ch5/obsact}(a-c) show  performance degradation of standard policies as increasing observation, action, and environmental noise are added at testing. We can see that policies are relatively sensitive to observation noise compared to the other types. In terms of domain-shift rather than noise, Figures~\ref{fig: ch5/obsact}(e-g) show that the performance \doublecheck{$\eta_{\Theta_{te}}$} 
of a standard model degrades rapidly as example environmental parameters (mass ratio, wind direction) are changed at testing, with the degradation rate depending on the factor being modified (e.g., greater mass-sensitivity than wind). Figure~\ref{fig: ch5/obsact}(d) summarizes the domain-shift performance as an average over increasing shift in all walker parameters (Table~\ref{table:env}: mass, wind, friction, gravity). In  this  particular environment,  TRPO  is usually the most robust algorithm.   
However, in the full evaluation over all tasks (Table~\ref{table: network1} in appendix) using testing AUC score (area under the curves in Fig~\ref{fig: ch5/obsact}), there is no consistent winner in algorithm robustness.

\cut{
\begin{table}[tb]
\centering
\caption{Improving Walker2d-PPO generalization to noisy testing by training with noise. Cols: Deterministic ($\sigma_{tr}=0.0$) vs noisy ($\sigma_{tr}=0.2$) training. Rows: Noise Types. Subrows: $\eta_{\sigma_{tr}}(\pi)$: Training return at noise level $\sigma_{tr}$. $\eta_{\sigma_{te}}(\pi)$: Testing return at noise $\sigma_{te}$. For simplicity we overload $\sigma$ to refer to $\Sigma$ in the Domain Shift setting, and use $\eta_\sigma(\pi)$ to indicate expected return $\eta_{p(\Theta;\Sigma)}(\pi)$.}\label{table: noise}
\begin{center}
\begin{small}
\begin{sc}
\input{ch4/table3.tex}
\end{sc}
\end{small}
\end{center}
\end{table}
}

 \begin{table}[tb]
 \centering
 \caption{Improving Walker2d-PPO generalization  by training with noise. (a) Compares training return in deterministic ($\sigma_{tr} = 0.0$) and noisy  ($\sigma_{tr} = 0.2$) conditions. (b) Compares the testing performance of different training conditions. 
 \cut{Cols: Deterministic ($\sigma_{tr}=0.0$) vs noisy ($\sigma_{tr}=0.2$) training. Rows: Noise Types. Subrows: $\eta_{\sigma_{tr}}(\pi)$: Training return at noise level $\sigma_{tr}$. $\eta_{\sigma_{te}}(\pi)$: Testing return at noise $\sigma_{te}$.}`MDL' and `DOM' refer to training or testing on multiple domains respectively. For simplicity we overload $\sigma$ to refer to $\Sigma$ in the multi-domain  setting, and use $\eta_\sigma(\pi)$ to indicate expected return $\eta_{p(\Theta;\Sigma)}(\pi)$.}\label{table: noise}
\vspace{-0.2in}
\hspace{-0.15in} 
\begin{center}

\begin{small}
\begin{sc}
\subtable[Training performance $\eta_{\sigma_{tr}}$ of deterministic vs. noisy training]{
\resizebox{1.0\columnwidth}{!}{
\begin{tabular*}{\linewidth}{c c c}
\toprule
Train with $\sigma_{tr} = 0.0$& \multicolumn{2}{c}{Train with $\sigma_{tr} = 0.2$} \\ 
\midrule
Return $\eta_{\sigma_{tr}}(\pi) $  & Noise Type & Return $\eta_{\sigma_{tr}}(\pi) $\\
\midrule
\multirow{4}{*}{$3128.6 \pm 402.3$}  & Obs. & $1424.0 \pm 472.9$ \\
&Act. & $2694.8\pm504.9$\\
&Env. & $ 3261.3 \pm 436.5$ \\
& MDL. & $2445.3 \pm 631.0$ \\
\bottomrule
\end{tabular*}} \label{table: noise1}}
 \end{sc}
 \end{small}

\vspace{-0.2in}
\hspace{-0.1in} 
\begin{small}
\begin{sc}
\subtable[Testing performance $\eta_{\sigma_{te}}$ of deterministic vs. noisy training]{
\resizebox{1.0\columnwidth}{!}{
\begin{tabular}{l c c c c c }
\toprule
\multicolumn{2}{l}{\multirow{2}{*}{Noise Type}} & \multirow{2}{*}{$\sigma_{te}$}  & Train with  \cut{($\sigma_{tr} = 0.0$)} & Train with\\
& & & $\sigma_{tr} = 0.0$ & $\sigma_{tr} = 0.2$ \\
\midrule
\multirow{3}{*}{Obs.}& \cut{\multirow{3}{*}{$\eta_{\sigma_{te}}(\pi)$}} & 0.0 & 3723.9$\pm$ 321.3  & 3682.9$\pm$415.6 \\
&& 0.2 & \textcolor{green}{2522.2 $\pm$593.4}  & \textcolor{green}{3036.1$\pm$576.3}\\
&& 0.4 & 937.8 $\pm$ 308.9 & 1303.8$\pm$374.5\\
\midrule
\multirow{3}{*}{Act.} &\cut{\multirow{3}{*}{$\eta_{\sigma_{te}}(\pi)$}}& 0.0 & 3678.8$\pm$355.5 &3596.5 $\pm$406.3\\
&& 0.2 & \textcolor{green}{2096.3$\pm$570.4} & \textcolor{green}{3044.0$\pm$525.6}\\
&& 0.4 & 919.7$\pm$249.5 & 1951.5$\pm$486.9\\
\midrule
\multirow{3}{*}{Env.} &\cut{\multirow{3}{*}{$\eta_{\sigma_{te}}(\pi)$}}& 0.0   &  3756.3$\pm$341.7 & 3761.8$\pm$415.1\\
&& 0.2 &  \textcolor{red}{3133.5$\pm$465.9} & \textcolor{red}{3225.4$\pm$450.0}\\
&& 0.4 & 2528.6$\pm$415.9	 & 2660.1 $\pm$ 436.1 \\
\midrule
\multirow{3}{*}{Dom.} &\cut{\multirow{3}{*}{$\eta_{\sigma_{te}}(\pi)$}}& 0.0   & 3725.1$\pm$343.7 & 3417.9$\pm$549.2\\
&& 0.2 & \textcolor{red}{3228.7$\pm$408.6} & \textcolor{red}{2985.2$\pm$486.4}\\
&& 0.4 & 2462.3$\pm$394.4 & 2468.2$\pm$341.9\\
\bottomrule
\end{tabular}} \label{table: noise2}}
\end{sc}
\end{small}
\end{center}
\vspace{-0.2in}
 \end{table}

\cut{
\begin{figure*}[t]
\centering
\subfigure[Observation Noise $\sigma^o$]
{\includegraphics[width=.49\columnwidth]{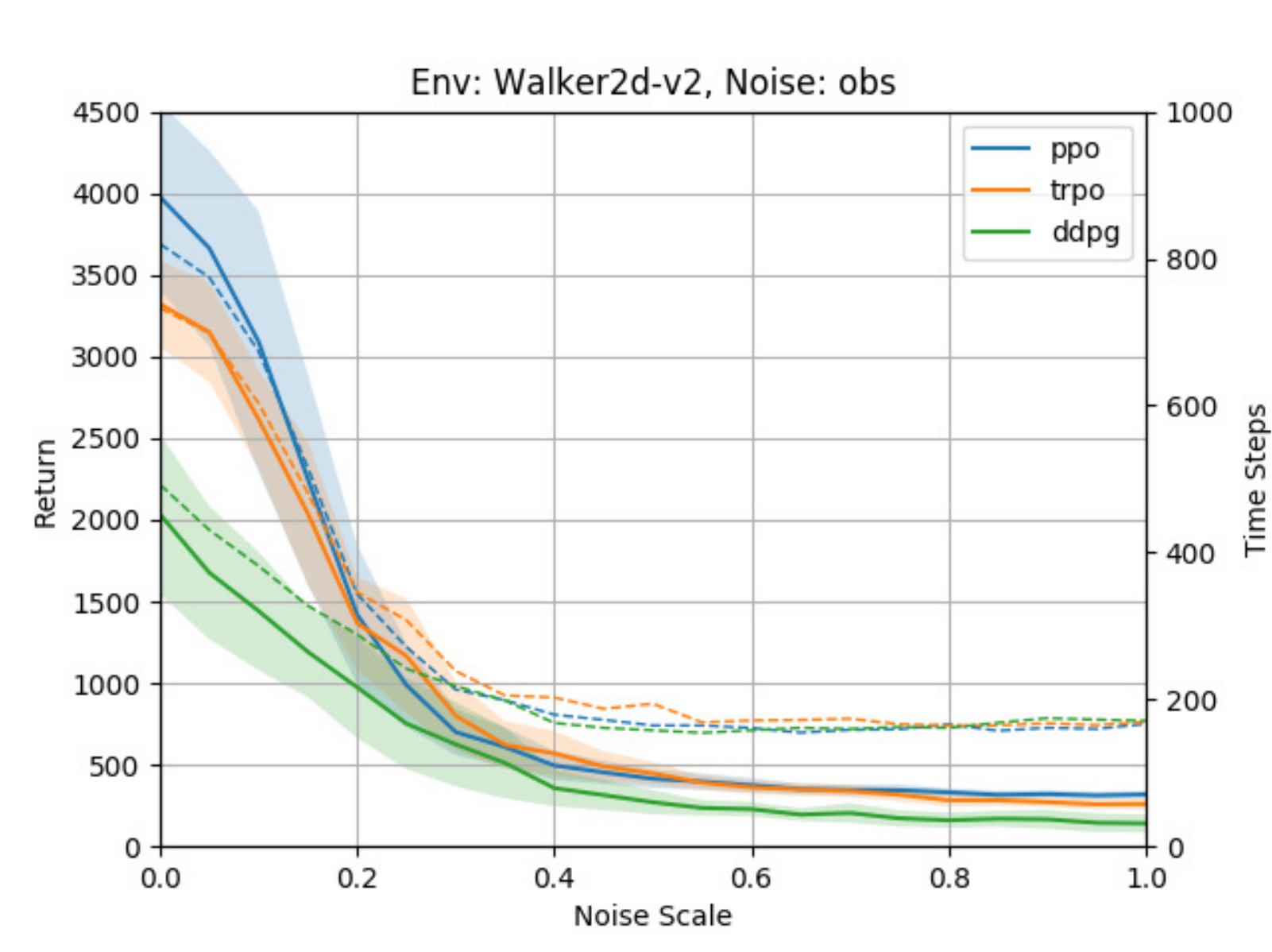}}
\subfigure[Action Noise  $\sigma^a$]
{\includegraphics[width=.49\columnwidth]{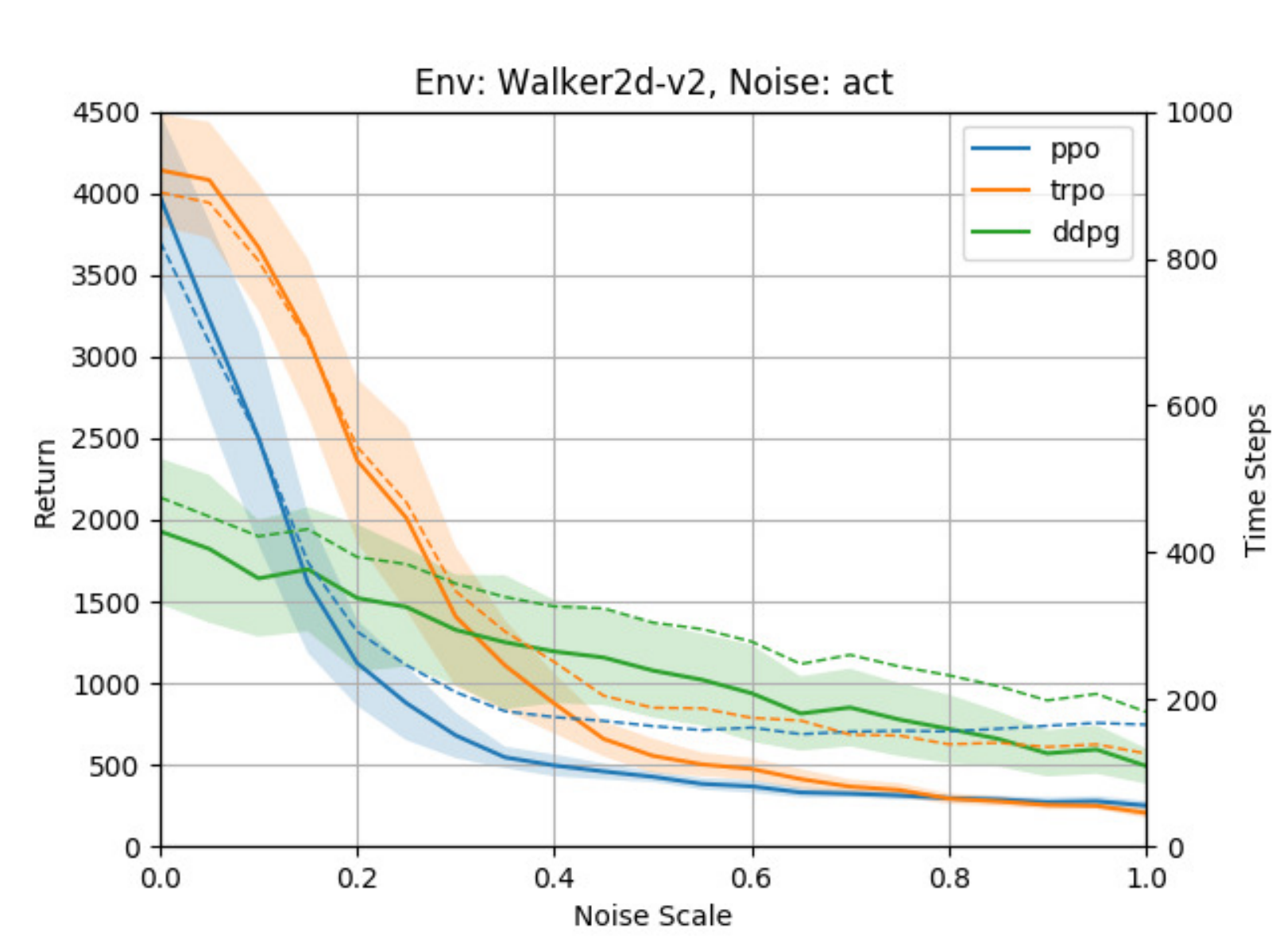}}
\subfigure[Environmental Noise $\sigma^e$]
{\includegraphics[width=.49\columnwidth]{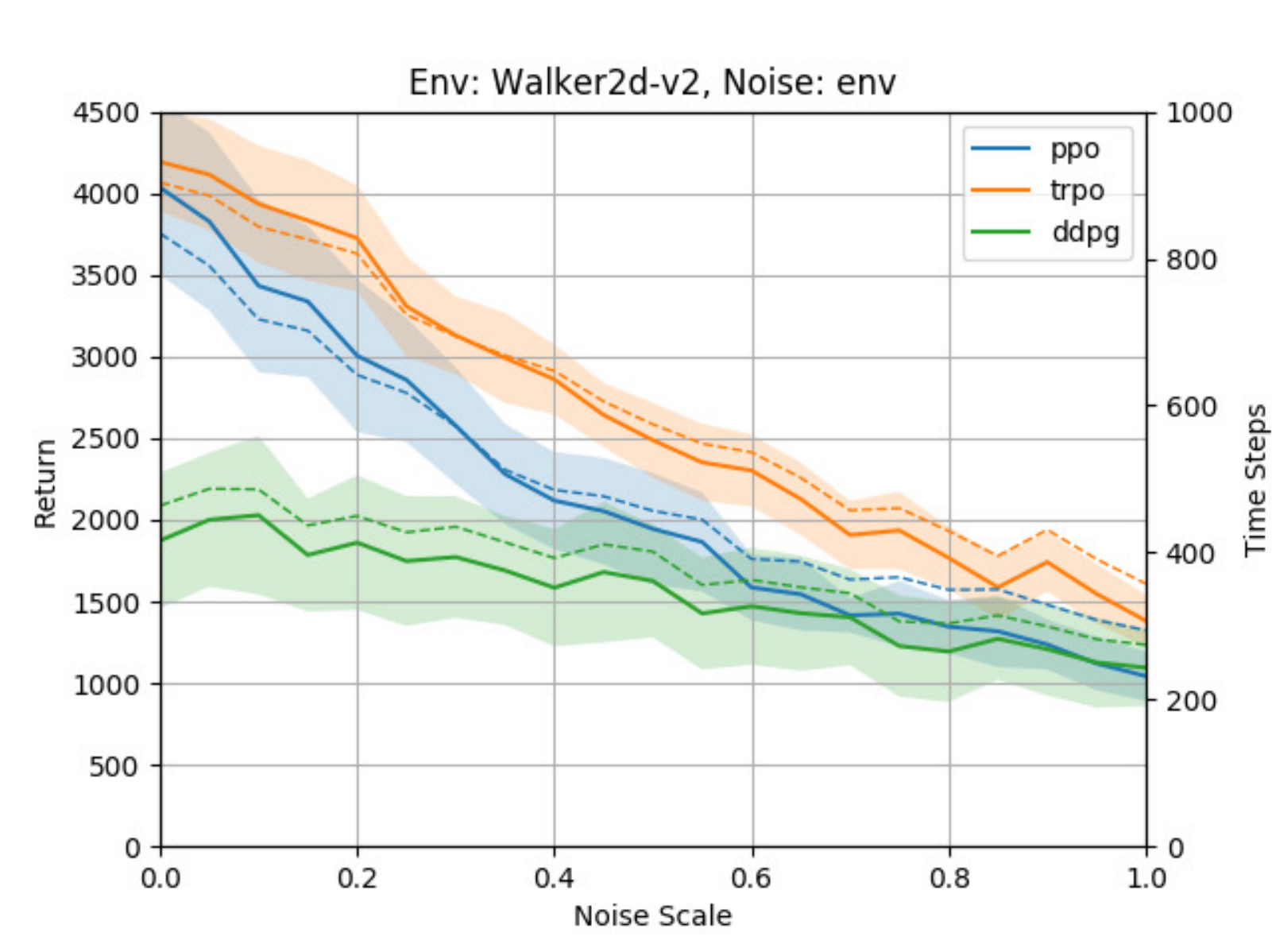}}
\subfigure[Domain Shift $\Sigma$]
{\includegraphics[width=.49\columnwidth]{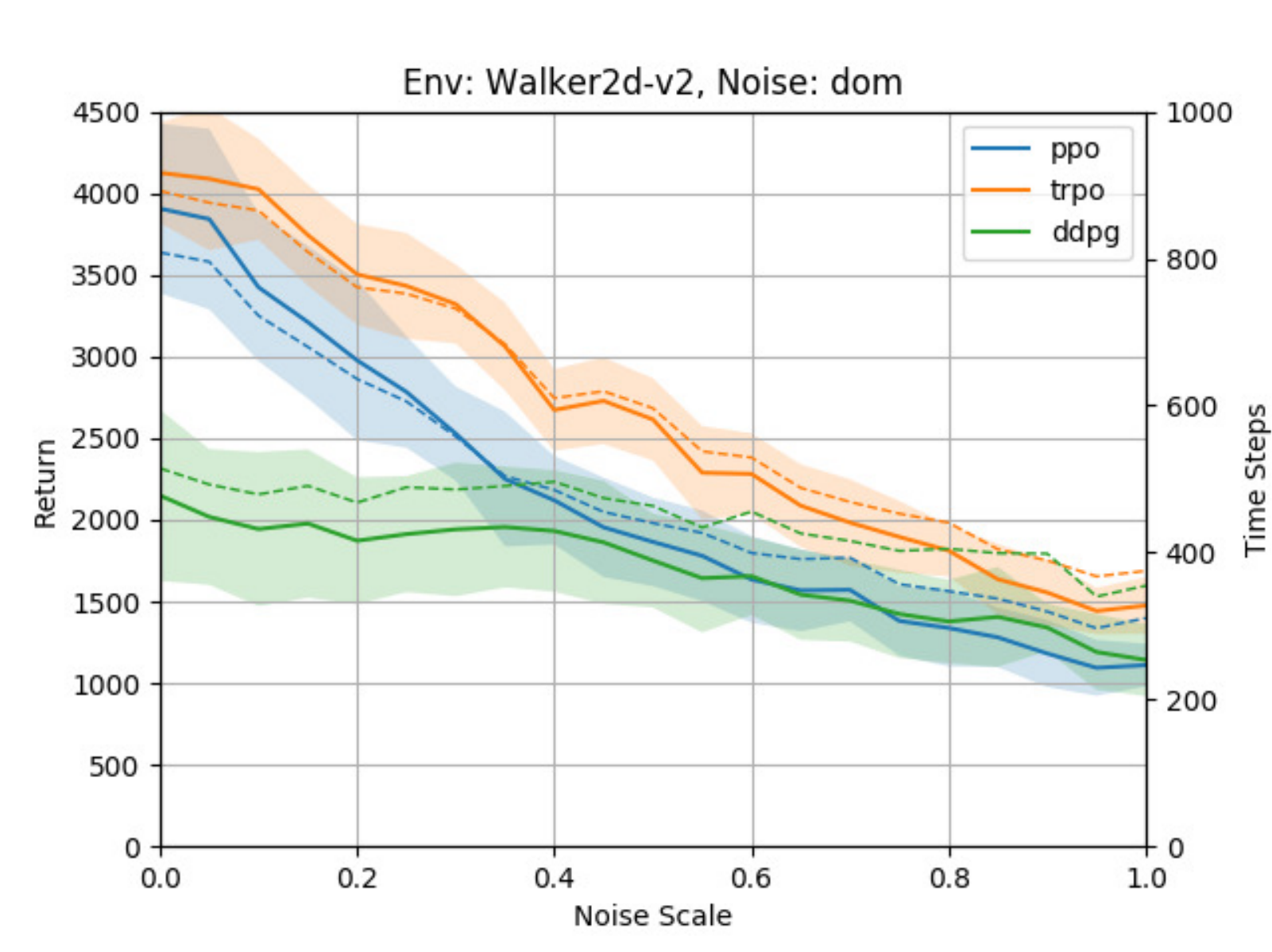}}
\vskip -0.1in
\centering
\subfigure[PPO]
{\includegraphics[width=.55\columnwidth]{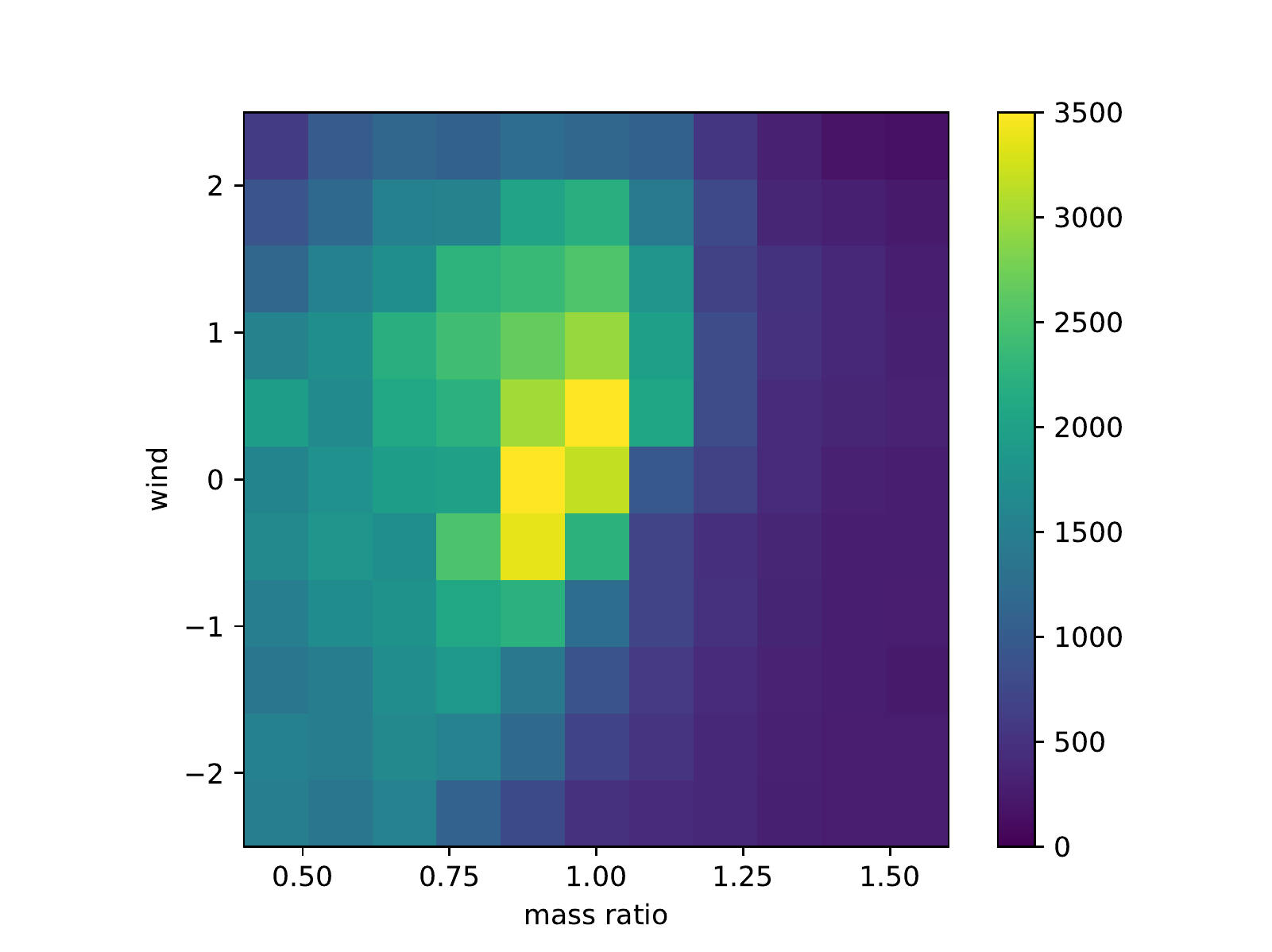}}
\subfigure[TRPO]
{\includegraphics[width=.55\columnwidth]{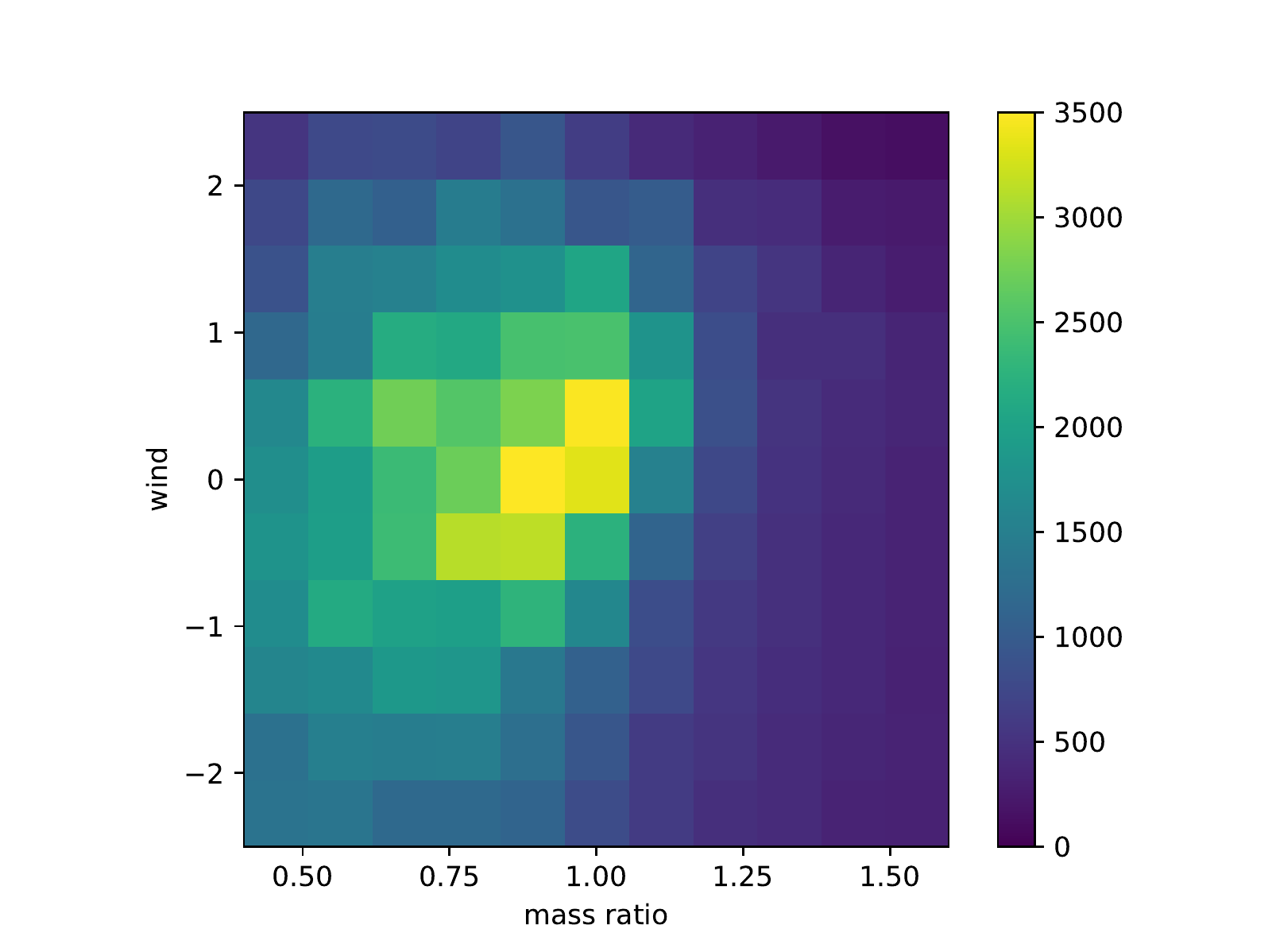}}
\subfigure[DDPG]
{\includegraphics[width=.55\columnwidth]{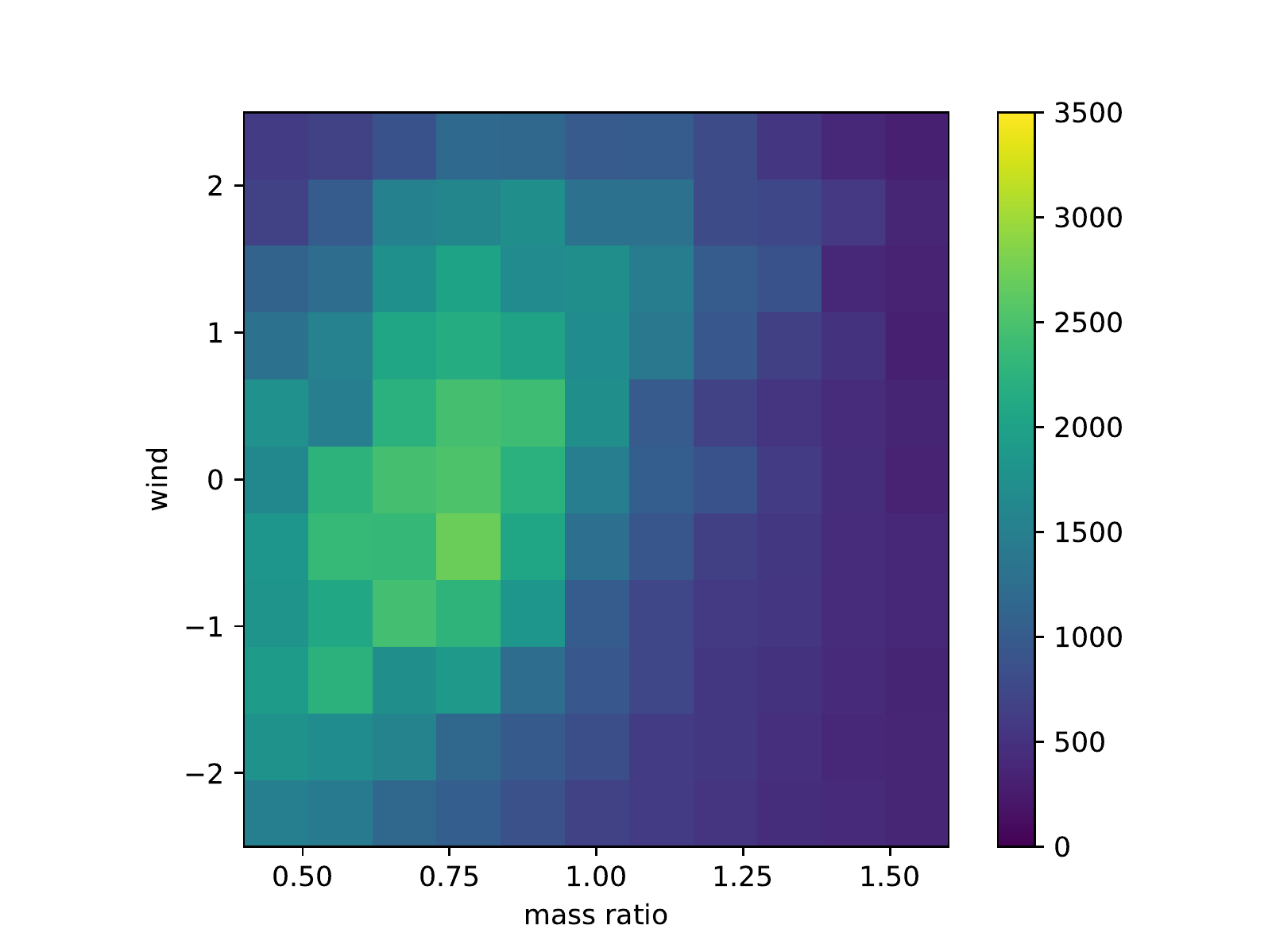}}
\vskip -0.1in
\caption{Generalization of standard continuous control policies for \textit{Walker2d-v2}. Top: Performance with varying testing noise \doublecheck{$\sigma^{(o,a,e)}$} and  environment variation $\Sigma$ scale. Bottom: Heatmaps illustrate policy performance over a grid of environmental domain-shifts. Each cell corresponds to a particular set of context parameters $\theta^{h}$ with training domain at $(0,0)$. Results are averaged over 12 random seeds.}
\label{fig: ch5/obsact}
\end{figure*}
}

\keypoint{Does modelling noise and variability in training improve generalisation?}

We saw above that performance degrades rapidly with noise. However, as discussed earlier, testing a deterministically trained  policy in a stochastic environment can be seen as a form of domain-shift ($\sigma_{tr}=0\to\sigma_{te}>0$). We therefore study if reducing this domain shift by adding noise and environment variation during training improves generalisation.

As a detailed example, we analyse PPO-trained \textit{Walker2d} in Table~\ref{table: noise}. To reduce domain shift between training and testing, we train the policies in noisy environment ($\sigma_{tr}=0.2$) to align with a testing condition, in comparison to training with default environment ($\sigma_{tr} =0$). We compare both training performance $\eta_{\sigma_{tr}}$ (Table~\ref{table: noise}a) and testing return $\eta_{\sigma_{te}}$ under multiple testing noise levels $\sigma_{te}$ (Table~\ref{table: noise}b).\hidden{Covering training in both the default and noisy environment designed to align with testing (\doublecheck{Table~\ref{table: noise}, columns $\sigma_{tr}=0$ and  $\sigma_{tr}=0.2$}).} The experiment considers both i.i.d Gaussian noise and training domain randomisation in preparation for testing on novel domains (denoted `Dom'). The expected testing return results (cf. Eq.~\eqref{eq:distRet}) are averaged over 12 training x 20 testing seeds.

    From the results in Table~\ref{table: noise}, we can see that (i) in each case, except environmental noise,  the (expected) training return is significantly lower when adding noise (Table~\ref{table: noise}a, compare cols)\hidden{(compare cols for $\eta_{\sigma_{tr}}$ rows)}, (ii) For observation and action noise, training with noise in preparation for testing with noise improves performance compared to the conventional deterministic training (Table~\ref{table: noise}b, compare green cols). (iii) However, there is not a clear benefit from removing the domain shift in this way if the testing scenario contains environment noise, or novel domains compared to training (Table~\ref{table: noise}b, compare red cols). (iv) Finally, we evaluate the impact of a domain-shift  $\sigma_{tr}=0.2\to\sigma_{te}=0.4$ corresponding to mis-specified noise strength. \cut{Finally we can ask what happens if we attempt to train in a stochastic environment, but mis-specify the noise strength compared to the true environment. For this we evaluate the domain shift $\sigma_{tr}=0.2\to\sigma_{te}=0.4$. } In this case we can see that while testing performance has generally degraded at $\sigma_{te}=0.4$ compared to $\sigma_{te}=0.2$, the degradation is ameliorated significantly (compared to deterministic training) in the case of action noise and observation noise.\cut{, slightly in the case of observation noise, and not at all for the others.} Welch's t-test ($p<0.05$) is used for all significance test.  

\begin{figure}[tb]
\centering
\subfigure[Noisy Training]
{\includegraphics[width=.99\columnwidth]{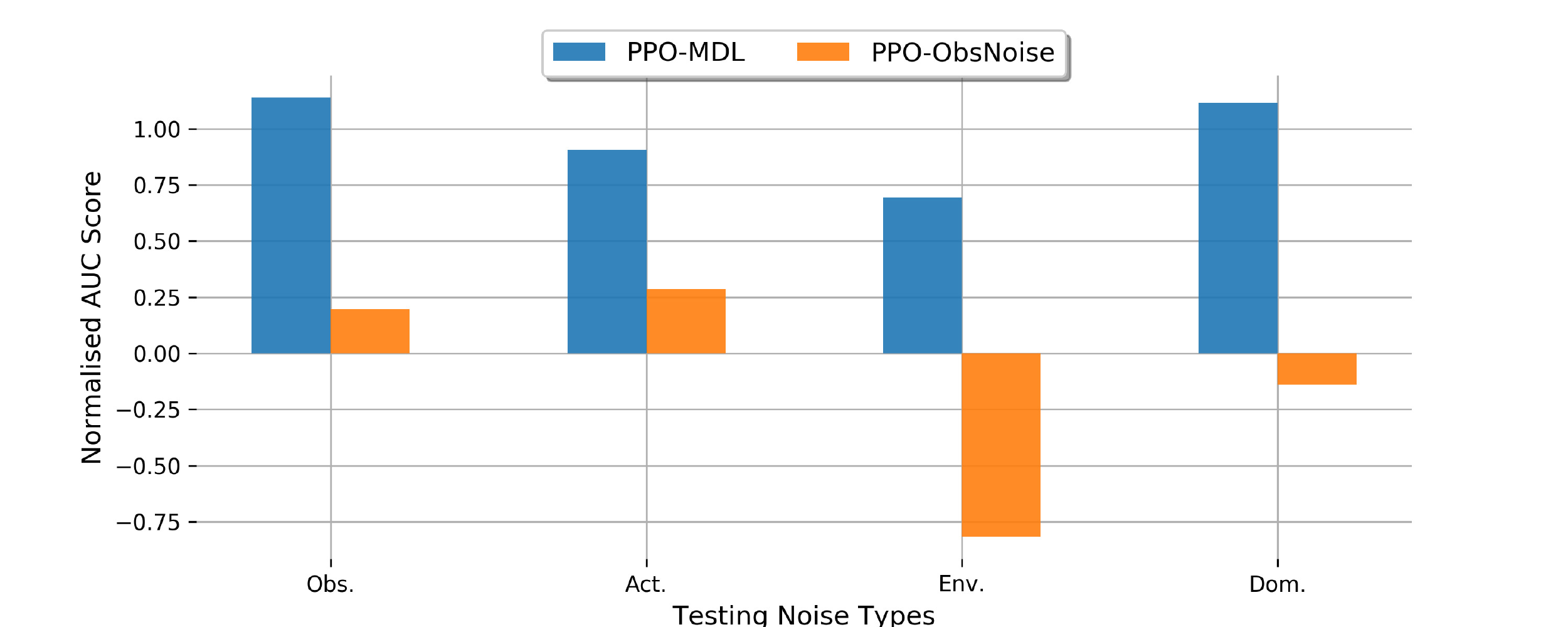}
\label{fig:ch5/block2}}
\subfigure[Training Techniques]
{\includegraphics[width=.99\columnwidth]{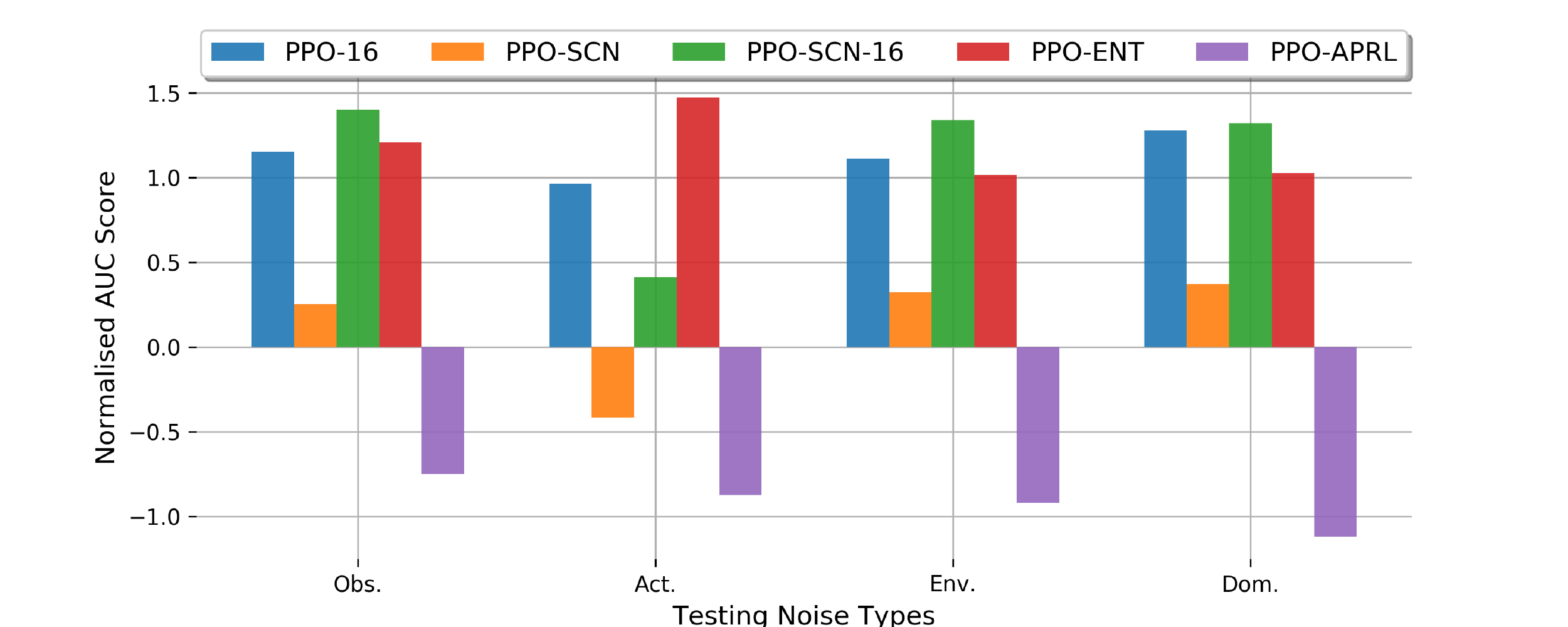}
\label{fig:ch5/block3}}
\vskip -0.1in
\caption{Change in normalised testing AUC score using various training settings compared to vanilla PPO training. (a) Impact of training with more stochastic environments. (b) Impact of architecture and algorithmic modifications. Results are averaged over all five training environments.}
\label{fig: ch5/block23}
\end{figure}

Figure~\ref{fig:ch5/block2} shows the impact of training with multiple domains or observation noise compared to training with deterministic environments, averaging over all five benchmark environments. Results are expressed as difference to vanilla PPO.  Detailed results across all tasks under the aggregate testing AUC metric (Eq.~\eqref{eq:AUC}) are visible in Table~\ref{table: network2} by comparing PPO (standard training) with PPO-MDL (Multi Domain Training), and PPO-Noisy Obs ($\sigma^o_{tr}=0.2$), where the green highlighted entries show when the noisy training improves on the PPO baseline. The results show positive influence in improving generalisation when training with noisy environments. However, adding noise during training can also raise the risk of failures in learning process for some tasks (especially training with noisy observations). Interestingly, there is some transferability across noise types. MDL training often improves robustness not only to novel domains at testing, but also i.i.d observation, action, and environment noise. Meanwhile, observation noise training improves robustness to action noise and MDL testing in \textit{HalfCheetah}. In summary, modeling uncertainty during training can help improve generalisation at testing, but better methods are still necessary particularly if uncertainty is misspecified.

\keypoint{What existing techniques improve generalisation?}

We next investigate if any of the methods discussed in Section~\ref{sec:algorithms} improve generalisation performance. We build on PPO due to being easier to integrate with the modifications (unlike, e.g., TRPO), and its good stability and training efficiency. Figure~\ref{fig:ch5/block3} illustrates the change in testing generalisation performance (normalised testing AUC score) compared to vanilla PPO, when using each training technique. Detailed expected testing AUC (Eq.~\eqref{eq:AUC}) results are summarised in Table~\ref{table: network3} in appendix.\hidden{in Table~\ref{table: network} summarise the full generalisation experiment over all algorithm variants.} We can see that: (i) The smaller PPO-16  often surpasses the classic PPO architecture with 64 hidden units each layer in generalisation, and similarly for SCN vs SCN-16. (ii) Entropy-regularised PPO usually surpasses vanilla PPO, sometimes by a large margin. (iii)  Adversarial PPO-APRL sometimes improves, but often also worsens vanilla PPO. (iv) The best performing model is either PPO-16, SCN-16, or PPO-Ent. Thus generalisation performance can be increased by reducing architecture size, or adding regularisers to reduce overfitting. However, there is no specific overfitting reduction strategy that works consistently across environments and noise types.

\cut{\begin{figure*}[h]
\centering
\subfigure[Walker2d-v2]
{\includegraphics[width=.66\columnwidth]{ch5/structure/Walker2d-v2_obs_2.eps}}
\subfigure[Hopper-v2]
{\includegraphics[width=.66\columnwidth]{ch5/structure/Hopper-v2_obs_2.eps}}
\subfigure[HalfCheetah-v2]
{\includegraphics[width=.66\columnwidth]{ch5/structure/HalfCheetah-v2_obs_2.eps}}
\vskip -0.1in
\subfigure[InvertedPendulum-v2]
{\includegraphics[width=.66\columnwidth]{ch5/structure/InvertedPendulum-v2_obs_2.eps}}
\subfigure[InvertedDoublePendulum-v2]
{\includegraphics[width=.66\columnwidth]{ch5/structure/InvertedDoublePendulum-v2_obs_2.eps}}
\vskip -0.1in
\caption{Curve of AUC score with respect to noise scale in five different environments. Performance are averaged over 12 randoms seeds. Different scales of noise are added to observation space.}
\label{fig: ch5/structure}
\end{figure*}}


\keypoint{Is training return a valid metric for performing model selection?}

In Deep RL research, training return is the standard evaluation metric for comparing learning algorithms and architectures. Given that ultimately we should care about testing return, this practice is based on the strong assumption that there is no overfitting and all distributions are identical during training and testing. However, we now know that overfitting does occur,  modelling errors between training and testing are unavoidable and that the real world is noisy \cite{sunderhauf2018limits,Koos2013transferability}. So it is important to ask what is the implication of this evaluation practice on the algorithms and architectures we determine to be `winners'.

\begin{figure}[t]
\centering
\includegraphics[width = .8\columnwidth]{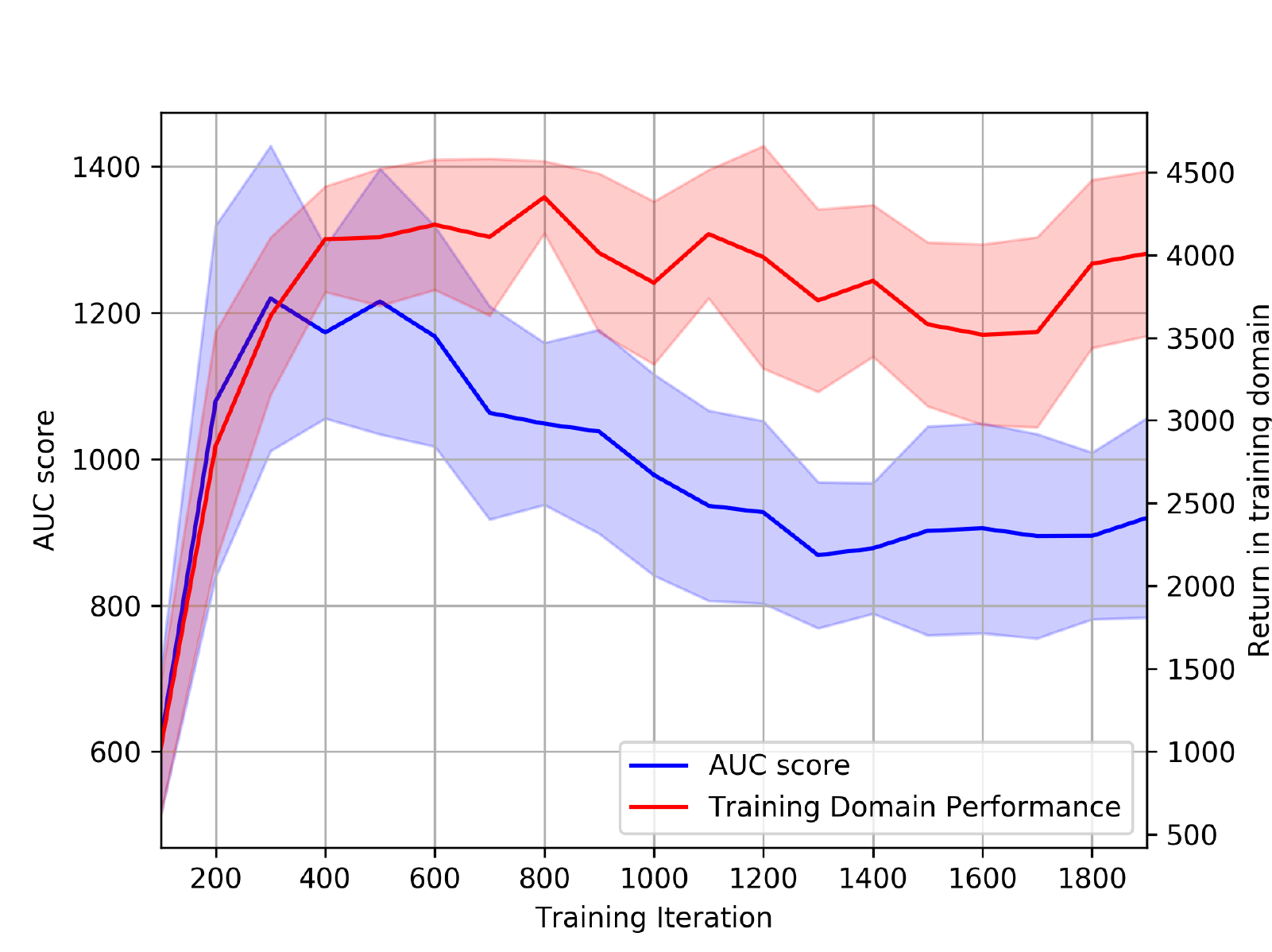}
\vskip -0.in
\caption{Comparison of testing AUC score and training return during PPO learning of \textit{Walker2d-v2} task. \cut{ and averaged over 12 trials}}
\label{fig: ch5/geneerror}
\end{figure}

\begin{figure}[t]
\centering
\subfigure[Walker2D - Action]
{\includegraphics[width=.49\columnwidth]{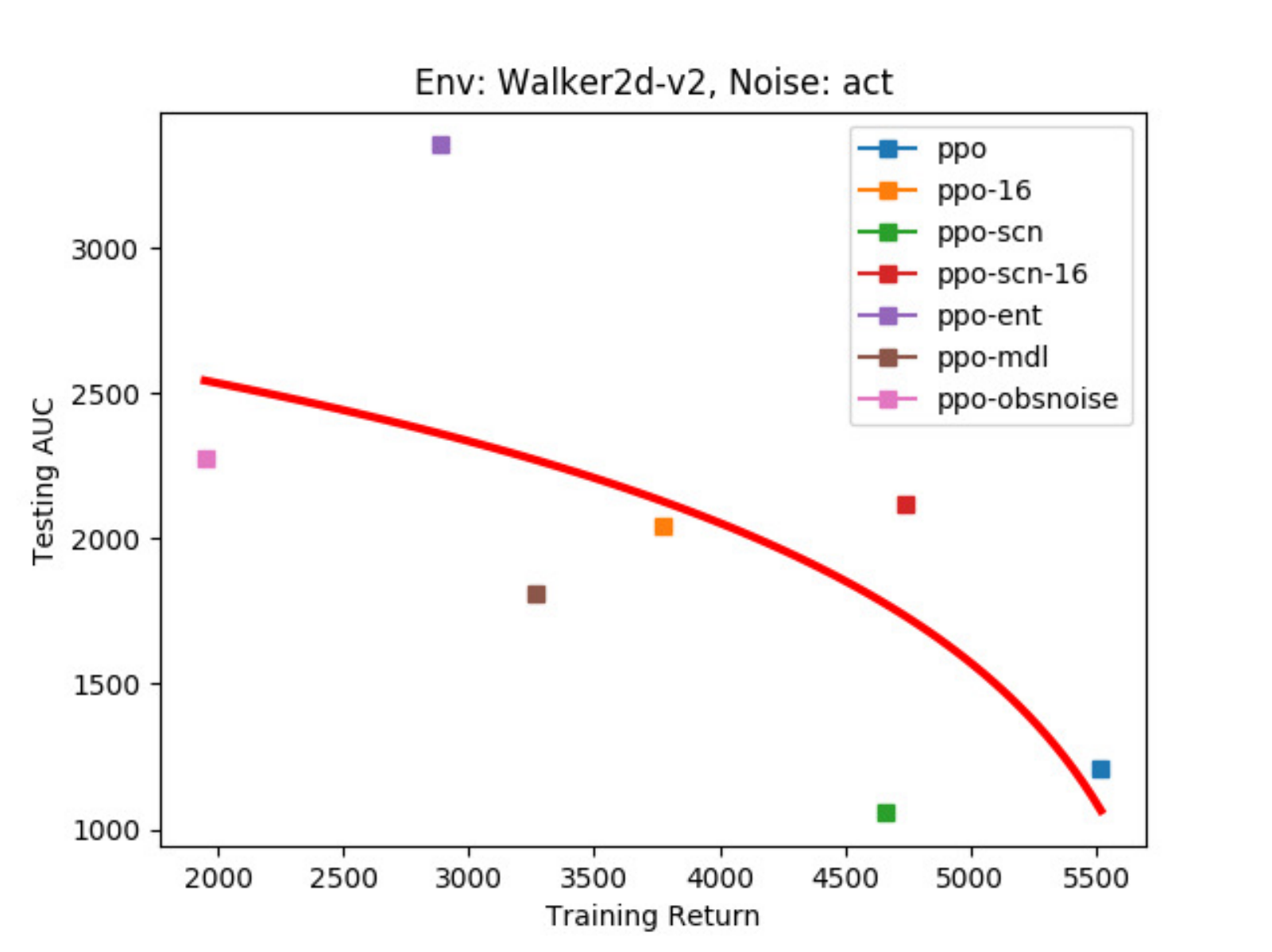}\label{fig: ch5/frontier1}}
\subfigure[HalfCheetah - Observation]
{\includegraphics[width=.49\columnwidth]{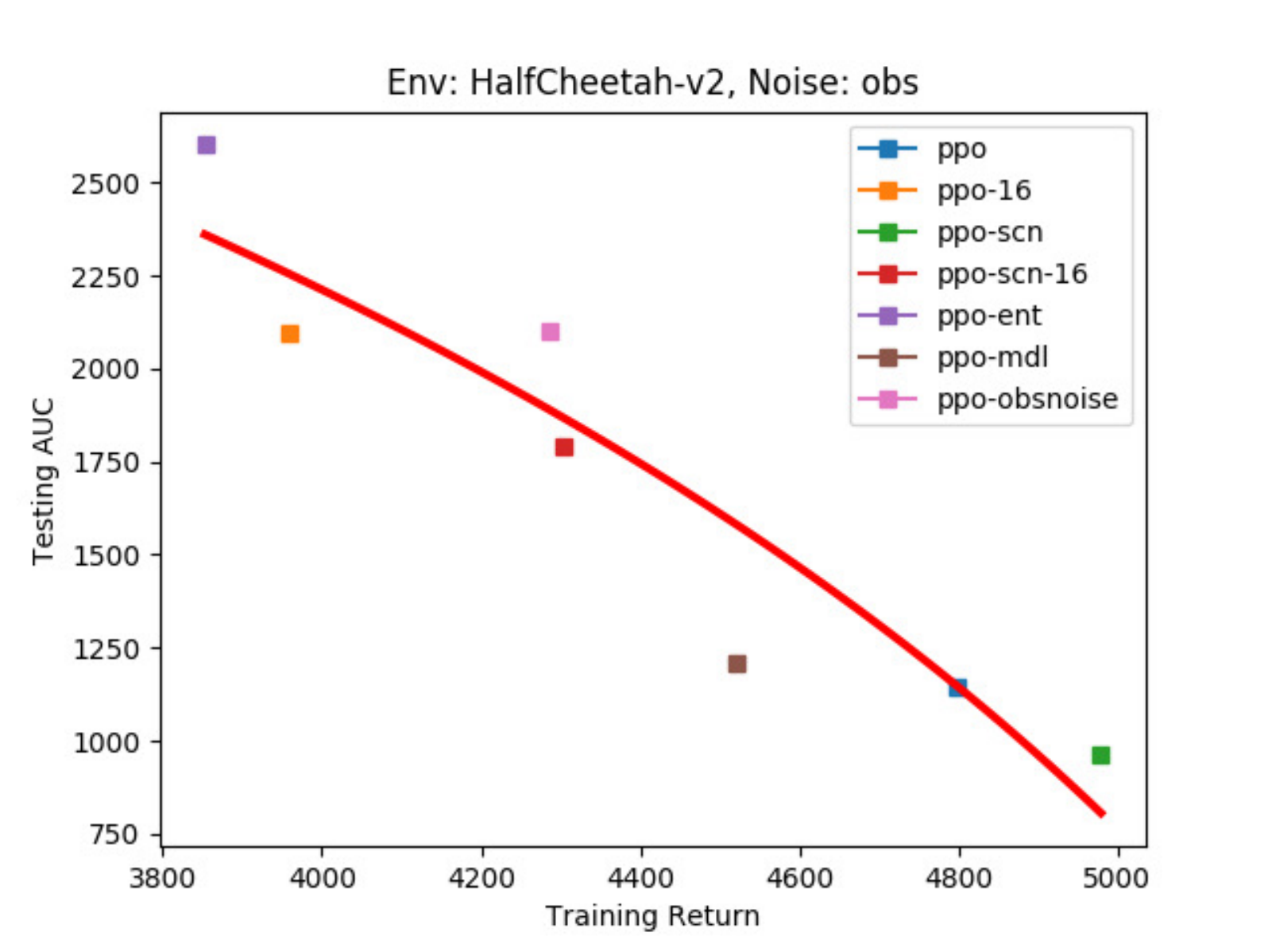}\label{fig: ch5/frontier2}}
\subfigure[Correlation Coef. between training returns and testing AUC]{
%
%
%
\begin{small}
\begin{sc}
\begin{tabular}{ l r r r r}
\toprule
& Obs. & Act.& Env. & Dom.\\
\midrule
Walker & -0.760& -0.722& -0.457 & -0.424 \\
Hopper & -0.203& -0.346& -0.473 & -0.415\\
HalfCheetah & -0.946& -0.843& -0.046 & -0.477\\
Pendulum & -0.132& -0.584& -0.665 & -0.630\\
D-Pendulum & -0.033& -0.732& ~0.025& ~0.324\\
\bottomrule\\
\end{tabular}
\end{sc}
\end{small}

\label{fig: ch5/frontier3}}
\vskip -0.1in
\caption{Training return does not reflect generalisation performance. (a,b): Clear Pareto frontiers exist in testing AUC vs training return across algorithms (dots). (c): Testing AUC and training return are generally negatively correlated.}
\label{fig: ch5/frontier}
\end{figure}

As an illustration, we first show how generalisation performance evolves during training. Figure~\ref{fig: ch5/geneerror} shows the testing  AUC score and training return as a function of PPO training iterations in the \textit{Walker2d-v2} environment. Training and testing performance initially improve in tandem, but overfitting occurs as learning continues. 
    To investigate the effect of this on algorithm choice, we fit a Pareto frontier to the testing AUC score vs training return of each method. Each algorithm is represented with the learned policy with best training performance among multiple random seeds. Similar results are obtained if using average performance across seeds. Figures~\ref{fig: ch5/frontier1},~\ref{fig: ch5/frontier2} show illustrative curves for  \textit{Walker2D}-ActNoise and \textit{HalfCheetah}-ObsNoise. The Pareto frontiers illustrate that it is hard to achieve good training and testing performance simultaneously. We further compute the correlation between testing AUC and training return for each task under each noise type in Figure~\ref{fig: ch5/frontier3}. \doublecheck{(Here the seven variants of the PPO algorithm in Fig~\ref{fig: ch5/frontier1} are the elements being correlated)}. Most environments and noise types have clear negative correlation. Thus if we follow standard practice of evaluating algorithms based on training performance, we will often pick  the least robust algorithm with worst generalisation. If generalisation is of interest, as it should be, then evaluations should use generalisation metrics such as the benchmarks proposed here. 

\section{Conclusion}
We contributed an analysis and set of benchmarks to investigate generalisation in deep RL.  The analysis showed that standard algorithms and architectures generalize poorly in the face of noise and environmental shift.  In particular, training and testing performance are often anti-correlated, so the standard practice of developing models with the aim of maximising training performance may be leading the community to produce less robust models. The results show that different off-the-shelf algorithms better address different aspects of generalisation performance, and various enhanced training strategies can also improve aspects of generalisation. However there is currently no generally good solution to all facets of generalisation, and new algorithms are needed.

\bibliography{ms}
\bibliographystyle{icml2019}
\clearpage

\section*{Appendix}
\cut{
\keypoint{Evaluation Environments}
\begin{table}[h]
\centering
\caption{Summary of evaluation environments and their environment factors that are included to generate shifts in transition model.}
\label{app table:env}
\begin{center}
\begin{small}
\resizebox{1.0\columnwidth}{!}{
\begin{tabular}{l l}
\toprule 
Task Name & Environment Factors\\
\midrule
InvertedPendulum-v2 & $m_{cart}$, $m_{pole}$\\
InvertedDoublePendulum-v2 & $m_{cart}$, $m_{pole_1}$, $m_{pole_2}$\\
Walker2d-v2 & $m_{body}$, $c_{wind}$, $c_{friction}$, $g$ \\
Hopper-v2 & $m_{body}$, $c_{wind}$, $c_{friction}$, $g$ \\
HalfCheetah-v2 & $m_{body}$, $c_{friction}$, $g$ \\
\bottomrule
\end{tabular}}
\end{small}
\end{center}
\end{table}
}

\cut{\keypoint{Entropy}
Given the success of the Entropy-regularisation strategy, we considered the hypothesis that the root cause of better generalisation is action entropy, and that the other methods such as smaller architectures work because smaller architectures that overfit less naturally produce higher-entropy policies. In this case \doublecheck{training?}-entropy could be an indicator of subsequent generalisation performance, without the need to evaluate testing environments.  To further investigate this we correlated policy entropy and AUC score in all environments. It happens that out of \doublecheck{N} tasks and conditions, X and Y had positive and negative correlation respectively between entropy and AUC. Therefore there is simpler relation between entropy and generalisation performance. \todo{Include/exclude? No space in main paper}}

\keypoint{Training 	Hyperparameters}

We use the implementation of basic RL algorithms (PPO, TRPO, DDPG) from OpenAI baselines codebase \cite{baselines}. The hyperparameters we use for each training algorithm are listed below:

$\bullet$ PPO

- Policy Network: (64, tanh, 64, tanh, Linear) + policy standard deviation variable \newline
- Value Network: (64, tanh, 64, tanh, Linear) \newline
- Normalised observations with running mean filter \newline
- Number of time steps per batch: 2048 \newline
- Number of optimiser epochs per iteration: 10 \newline
- Size of optimiser minibatches: 64 \newline
- Optimiser learning rate: $3e-4$ \newline
- Generalised Advantage Estimator (GAE) factor $\lambda = 0.98$ \newline
- Discount factor $\gamma = 0.99$ \newline
- Cliprange parameter: 0.2\newline
- Number of total training steps: $4e6$

$\bullet$ TRPO

- Policy Network: (64, tanh, 64, tanh, Linear) + policy standard deviation variable \newline
- Value Network: (64, tanh, 64, tanh, Linear) \newline
- Normalised observations with running mean filter \newline
- Number of time steps per batch: 1024 \newline
- Maximum KL divergence: $0.01$ \newline
- Conjugate gradient iterations: 10 \newline 
- CG damping factor: $0.1$ \newline 
- Generalised Advantage Estimator (GAE) factor $\lambda = 0.98$ \newline
- Discount factor $\gamma = 0.99$ \newline
- Value network update epochs per iteration: 5 \newline
- Value network learning rate: $1e-3$ \newline
- Number of total training steps: $4e6$

$\bullet$ DDPG

- Actor Network: (64, relu, 64, relu, tanh) \newline
- Critic Network: (64, relu, 64, relu, Linear) \newline
- Normalised observations with running mean filter \newline
- Noise type: OU-Noise 0.2 \newline
- Learning rates: actor LR: $1e-4$, critic LR: $1e-3$ \newline 
- L2 normalisation coeff: 0.01 \newline
- Batch size: 64 \newline
- Discount factor $\gamma = 0.99$ \newline
- Soft target update $\tau = 0.01$ \newline
- Reward Scale: 1.0\newline
- Number of total training steps: $4e6$

\cut{\keypoint{Temporary}
\begin{table*}[h!]
\centering
\caption{Walked2d task: Training with noise in preparation for testing with noise. \todo{Missing clear notation for $\sigma$ in domain randomisation.}}
\label{table: noise}
\vskip 0.15in
\begin{center}
\begin{small}
\begin{sc}
\begin{tabular}{l c c c c c c}
\toprule
\multirow{2}{*}{Noise Type}& \multicolumn{3}{c}{Training with Default Env ($\sigma_{tr} = 0.0s$)} &\multicolumn{3}{c}{Training in Noisy Environments ($\sigma_{tr} = 0.2$)}\\
& Training Return & $ \sigma_{te}$ &$\eta_{\sigma_{te}}(\pi)$ & Training Return & $ \sigma_{te}$ & $\eta_{\sigma_{te}}(\pi)$ \\
\midrule
\multirow{3}{*}{Observation} & \multirow{3}{*}{3128.6 $\pm$ 402.3} & 0.0 & 3723.9$\pm$ 321.3 & \multirow{3}{*}{1424.0$\pm$472.9} & 0.0 & 3682.9$\pm$415.6 \\
& & 0.2 & 2522. 2 $\pm$593.4  & & 0.2 & 3036.1$\pm$576.3\\
& & 0.4 & 937.8 $\pm$ 308.9 & & 0.4 & 1303.8$\pm$374.5\\
\midrule
\multirow{3}{*}{Action} & \multirow{3}{*}{3128.6 $\pm$ 402.3} & 0.0 &3678.8$\pm$355.5  & \multirow{3}{*}{2694.8$\pm$504.9} & 0.0 & 3596.5 $\pm$406.3 \\
& & 0.2 & 2096.3$\pm$570.4 & & 0.2 & 3044.0$\pm$525.6\\
& & 0.4 & 919.7$\pm$249.5 & & 0.4 &1951.5$\pm$486.9\\
\midrule
\multirow{3}{*}{Environment} & \multirow{3}{*}{3128.6 $\pm$ 402.3} & 0.0 & 3756.3$\pm$341.7 & \multirow{3}{*}{3261.3$\pm$436.5} & 0.0 & 3761.8$\pm$415.1 \\
& & 0.2 &  3133.5$\pm$465.9 & & 0.2 & 3225.4$\pm$450.0\\
& & 0.4 & 2528.6$\pm$415.9	 & & 0.4 & 2660.1 $\pm$ 436.1 \\
\midrule
\multirow{3}{*}{MultiDomain} & \multirow{3}{*}{3128.6 $\pm$ 402.3} & 0.0 & 3725.1$\pm$343.7 & \multirow{3}{*}{2445.3$\pm$631.0} & 0.0 & 3417.9$\pm$549.2\\
& & 0.2 & 3228.7$\pm$408.6 & & 0.2 & 2985.2$\pm$486.4\\
& & 0.4 & 2462.3$\pm$394.4 & & 0.4 & 2468.2$\pm$341.9\\
\bottomrule
\end{tabular}
\end{sc}
\end{small}
\end{center}
\end{table*}}

\begin{table*}[t!]
\tiny
\centering
\caption{Generalisation performance of basic Deep RL learners in terms of AUC scores. All settings are evaluated with 5 environments and 4 noise types in each environment. All results are averaged over 12 random seeds. Blue: Winning algorithm among the basic PPO, TRPO and DDPG baselines. Welch's t-test ($p<0.05$) is used for significance testing.}
\label{table: network1}
\vskip 0.15in
\begin{center}
\begin{small}
\begin{sc}

\begin{tabular}{l l c c c HHZ}
\toprule
\multicolumn{2}{l}{env. Name} & PPO & TRPO & DDPG & PPO-MDL & PPO-ObsNoise  \\
\midrule
Walker2d & obs. & \hblue{$\phantom{0}966.2 \pm 308.2$} & $\phantom{0}895.4 \pm 229.3$ & $\phantom{0}544.3 \pm 204.4$ & $\hgreen{1114.3\pm378.2}$ & $\hgreen{1455.8\pm388.9}$\\
& act. & $\phantom{0}845.5 \pm 251.4 $& \hblue{$1295.5 \pm 310.2$} &  $1115.2 \pm 531.1 $ &  \hgreen{$1082.1\pm349.6$} & $\hgreen{\mathbf{2078.3\pm522.8}}$\\
& env. & $2141.1 \pm 558.5$ & \hblue{$2653.6 \pm 402.3$} & $1550.5\pm 671.4$  & $2232.5\pm609.4$& $2301.8\pm521.0$\\
& dom. & $2114.3\pm567.2 $& \hblue{$2648.4 \pm 417.5$} & $1929.4\pm 839.8$ & $2232.9\pm601.8$ & $2313.7\pm521.9$ \\
\midrule
Hopper & obs. & $\phantom{0}592.6\pm 244.4$  & \hblue{$\phantom{0}643.8\pm 107.2$} &  $\phantom{0}352.7\pm245.5$  & $\hgreen{\phantom{0}733.5\pm143.4}$  & $\phantom{0}582.2\pm318.9$ \\
& act. & $\phantom{0} 609.0\pm167.6$ & \hblue{$\phantom{0}683.0\pm121.8$} & $\phantom{0}582.3\pm 163.9$ & $\hgreen{\phantom{0}905.8\pm143.4}$ & $\phantom{0}683.8\pm351.7$ \\
& env. & $1241.6\pm 438.7$ & \hblue{$1638.7\pm383.7$} &  $1042.2\pm337.9$& $\hgreen{1935.8\pm428.4}$ & $1164.9\pm804.3$\\
& dom. & \hblue{$1640.0\pm507.4$} & \hblue{$1622.6\pm 374.2$} & $1073.5\pm 358.3$ &  $\hgreen{1939.8\pm425.0}$ & $1167.8\pm813.6$ \\
\midrule
HalfCheetah & obs. & $1030.4\pm 133.9$ & $1045.9\pm \phantom{0} 444.6$ &  \hblue{$1152.7\pm361.0$} & $\hgreen{\mathbf{1629.6\pm\phantom{0}585.5}}$ & $\hgreen{\mathbf{1787.3\pm\phantom{0}562.8}}$ \\
& act. & $\phantom{0}773.1\pm177.5$ & $\phantom{0}906.1\pm   \phantom{0} 530.3$ & \hblue{$\phantom{0}943.6\pm312.8$}  & $\hgreen{1307.9\pm\phantom{0}498.9}$ & $\hgreen{\mathbf{1882.6\pm\phantom{0}627.2}}$\\
& env. & $1989.0\pm 658.2 $ & \hblue{$2322.0 \pm 1235.3$} & $1856.4 \pm 481.0$ & $1935.8\pm \phantom{0} 428.4$ & $1164.9\pm \phantom{0} 804.3$  \\
& dom. & $1871.0\pm583.5$ & \hblue{$2264.7\pm1191.2$} & $1963.9\pm451.7$ & $\hgreen{\mathbf{3748.2\pm1550.7}}$ & $\hgreen{\mathbf{3588.5\pm1020.7}}$\\
\midrule
Pendulum & obs. & $129.5\pm\phantom{0}32.5$ & \hblue{$182.8\pm 139.9$} & $\phantom{0}94.9\pm\phantom{0}33.1$  & $\hgreen{162.3\pm39.9}$ & $\phantom{0}54.1\pm\phantom{0}25.0$ \\
& act. & \hblue{$362.7\pm\phantom{0}43.7$} & $201.0\pm112.8$&$245.3\pm\phantom{0}85.3$ & $\hgreen{393.0\pm16.2}$ & $250.8\pm109.4$\\
& env. & \hblue{$995.4\pm\phantom{0}\phantom{0}8.0$} & $678.9\pm371.3$ & $607.2\pm280.0$  & $\hgreen{\mathbf{999.1\pm\phantom{0}2.4}}$ & $774.5\pm382.4$\\
& dom. & \hblue{$942.2\pm177.6$} & $676.1\pm370.9 $& $609.2\pm281.3$ &$\hgreen{\mathbf{999.3\pm\phantom{0}1.8}}$ & $775.0\pm381.8$ \\
\midrule
D-Pendulum & obs. & \hblue{$\phantom{0}923.3\pm\phantom{0}\phantom{0}61.5$} & $\phantom{0}715.1\pm\phantom{0}215.0$& $\phantom{0}525.4\pm257.9$  &$\hgreen{\mathbf{1071.8\pm\phantom{0}87.1}}$ & $\phantom{0}160.8\pm\phantom{0}13.9$ \\
& act. & \hblue{$2110.7\pm\phantom{0}\phantom{0}74.6$}& $1280.9\pm\phantom{0}286.4$ & $\phantom{0}962.4\pm178.7$  & $\hgreen{\mathbf{2161.5\pm\phantom{0}36.0}}$ & $\phantom{0}301.2\pm\phantom{0}64.6$ \\
& env. & \hblue{$7494.9\pm1189.1$} & $3023.9\pm\phantom{0}753.5$ & $2377.2\pm831.9$ & $\hgreen{\mathbf{8183.7\pm144.5}}$ & $\phantom{0}312.8\pm\phantom{0}50.5$ \\
& dom. & \hblue{$7013.8 \pm\phantom{0} 850.6$} & $6103.9\pm1530.1$ & $2215.8\pm 915.6$  &$\hgreen{\mathbf{8174.2\pm178.2}}$&$1178.9\pm688.1$ \\
\bottomrule
\end{tabular}
\end{sc}
\end{small}
\end{center}
\vskip -0.1in
\end{table*}

\begin{table*}[t!]
\tiny
\centering
\caption{Generalisation performance of training with noisy environments in terms of AUC score. All settings are evaluated with 5 environments and 4 noise types in each environment. All performance are averaged over 12 random seeds. Green: Noisy training settings that improve on corresponding PPO baseline. Welch's t-test ($p<0.05$) is used for significant testing.}
\label{table: network2}
\vskip 0.15in
\begin{center}
\begin{small}
\begin{sc}

\begin{tabular}{l l c H H c c}
\toprule
\multicolumn{2}{l}{env. Name} & PPO & TRPO & DDPG & PPO-MDL & PPO-ObsNoise  \\
\midrule
Walker2d & obs. & $\phantom{0}966.2 \pm 308.2$ & $\phantom{0}895.4 \pm 229.3$ & $\phantom{0}544.3 \pm 204.4$ & $\hgreen{1114.3\pm378.2}$ & $\hgreen{1455.8\pm388.9}$\\
& act. & $\phantom{0}845.5 \pm 251.4 $& \hblue{$1295.5 \pm 310.2$} &  $1115.2 \pm 531.1 $ &  \hgreen{$1082.1\pm349.6$} & $\hgreen{ {2078.3\pm522.8}}$\\
& env. & $2141.1 \pm 558.5$ & \hblue{$2653.6 \pm 402.3$} & $1550.5\pm 671.4$  & $2232.5\pm609.4$& $2301.8\pm521.0$\\
& dom. & $2114.3\pm567.2 $& \hblue{$2648.4 \pm 417.5$} & $1929.4\pm 839.8$ & $2232.9\pm601.8$ & $2313.7\pm521.9$ \\
\midrule
Hopper & obs. & $\phantom{0}592.6\pm 244.4$  & \hblue{$\phantom{0}643.8\pm 107.2$} &  $\phantom{0}352.7\pm245.5$  & $\hgreen{\phantom{0}733.5\pm143.4}$  & $\phantom{0}582.2\pm318.9$ \\
& act. & $\phantom{0} 609.0\pm167.6$ & \hblue{$\phantom{0}683.0\pm121.8$} & $\phantom{0}582.3\pm 163.9$ & $\hgreen{\phantom{0}905.8\pm143.4}$ & $\phantom{0}683.8\pm351.7$ \\
& env. & $1241.6\pm 438.7$ & \hblue{$1638.7\pm383.7$} &  $1042.2\pm337.9$& $\hgreen{1935.8\pm428.4}$ & $1164.9\pm804.3$\\
& dom. & $1640.0\pm507.4$ & \hblue{$1622.6\pm 374.2$} & $1073.5\pm 358.3$ &  $\hgreen{1939.8\pm425.0}$ & $1167.8\pm813.6$ \\
\midrule
HalfCheetah & obs. & $1030.4\pm 133.9$ & $1045.9\pm \phantom{0} 444.6$ &  \hblue{$1152.7\pm361.0$} & $\hgreen{ {1629.6\pm\phantom{0}585.5}}$ & $\hgreen{ {1787.3\pm\phantom{0}562.8}}$ \\
& act. & $\phantom{0}773.1\pm177.5$ & $\phantom{0}906.1\pm   \phantom{0} 530.3$ & \hblue{$\phantom{0}943.6\pm312.8$}  & $\hgreen{1307.9\pm\phantom{0}498.9}$ & $\hgreen{ {1882.6\pm\phantom{0}627.2}}$\\
& env. & $1989.0\pm 658.2 $ & \hblue{$2322.0 \pm 1235.3$} & $1856.4 \pm 481.0$ & $1935.8\pm \phantom{0} 428.4$ & $1164.9\pm \phantom{0} 804.3$  \\
& dom. & $1871.0\pm583.5$ & \hblue{$2264.7\pm1191.2$} & $1963.9\pm451.7$ & $\hgreen{ {3748.2\pm1550.7}}$ & $\hgreen{ {3588.5\pm1020.7}}$\\
\midrule
Pendulum & obs. & $129.5\pm\phantom{0}32.5$ & \hblue{$ {182.8\pm 139.9}$} & $\phantom{0}94.9\pm\phantom{0}33.1$  & $\hgreen{162.3\pm39.9}$ & $\phantom{0}54.1\pm\phantom{0}25.0$ \\
& act. & $362.7\pm\phantom{0}43.7$ & $201.0\pm112.8$&$245.3\pm\phantom{0}85.3$ & $\hgreen{ {393.0\pm16.2}}$ & $250.8\pm109.4$\\
& env. & $995.4\pm\phantom{0}\phantom{0}8.0$ & $678.9\pm371.3$ & $607.2\pm280.0$  & $\hgreen{ {999.1\pm\phantom{0}2.4}}$ & $774.5\pm382.4$\\
& dom. & $942.2\pm177.6$ & $676.1\pm370.9 $& $609.2\pm281.3$ &$\hgreen{ {999.3\pm\phantom{0}1.8}}$ & $775.0\pm381.8$ \\
\midrule
D-Pendulum & obs. & $\phantom{0}923.3\pm\phantom{0}\phantom{0}61.5$ & $\phantom{0}715.1\pm\phantom{0}215.0$& $\phantom{0}525.4\pm257.9$  &$\hgreen{ {1071.8\pm\phantom{0}87.1}}$ & $\phantom{0}160.8\pm\phantom{0}13.9$ \\
& act. & $2110.7\pm\phantom{0}\phantom{0}74.6$& $1280.9\pm\phantom{0}286.4$ & $\phantom{0}962.4\pm178.7$  & $\hgreen{ {2161.5\pm\phantom{0}36.0}}$ & $\phantom{0}301.2\pm\phantom{0}64.6$ \\
& env. & $7494.9\pm1189.1$ & $3023.9\pm\phantom{0}753.5$ & $2377.2\pm831.9$ & $\hgreen{ {8183.7\pm144.5}}$ & $\phantom{0}312.8\pm\phantom{0}50.5$ \\
& dom. & $7013.8 \pm\phantom{0} 850.6$ & $6103.9\pm1530.1$ & $2215.8\pm 915.6$  &$\hgreen{ {8174.2\pm178.2}}$&$1178.9\pm688.1$ \\
\bottomrule
\end{tabular}

\end{sc}
\end{small}
\end{center}
\vskip -0.1in
\end{table*}

\begin{sidewaystable*}[t!]
\tiny
\centering
\caption{Generalisation performance of different algorithms and architectures in terms of AUC score. All settings are evaluated with 5 environments and 4 noise types in each environment. All performance are averaged over 12 random seeds. The training settings that give top-$2$ performance within a test setting are highlighted in boldface. Welch's t-test ($p<0.05$) is used for significant testing. }
\label{table: network3}
\vskip 0.15in
\begin{center}
\begin{small}
\begin{sc}
\hspace{-0.5in}
\begin{tabular}{l l c c c c c c}
\toprule

\multicolumn{2}{l}{env. Name} & PPO & PPO-16 & PPO-SCN & PPO-SCN-16  & PPO-Ent & PPO-APRL \\
\midrule
Walker2d & obs. & $\phantom{0}966.2\pm 308.2$ & $ 1243.8 \pm 370.6$ & $1035.6\pm223.9$ & $\mathbf{1700.5\pm401.3}$&  $\mathbf{ 2442.6\pm \phantom{0}723.6 } $ & $\phantom{0}992.6\pm243.9$ \\
& act. & $\phantom{0}845.5\pm251.4$ & $ 1181.6\pm 418.2$ &$ \phantom{0}944.8\pm299.1$ & $\mathbf{1677.2\pm464.6}$ & $\mathbf{2843.9\pm 1075.6}$ & $\phantom{0}862.4\pm185.7$\\
& env. & $2141.1 \pm 558.5$& $ 2337.4 \pm 633.7$ &  $2180.5\pm458.9$ & $\mathbf{3050.9\pm805.8}$ & $\mathbf{3383.3 \pm \phantom{0} 725.8}$ & $2038.3\pm489.0$\\
& dom. & $2114.3 \pm 567.2$ & $ 2315.8\pm 618.4$ & $2166.6\pm447.6$ & $\mathbf{3039.5\pm810.8}$ & $\mathbf{3271.8 \pm \phantom{0} 740.4}$ & $1086.3\pm323.3$\\
\midrule
Hopper & obs. & $\phantom{0}592.6\pm 244.4$ & $\mathbf{\phantom{0}936.5\pm302.4}$ &  $\phantom{0}576.1\pm175.9$ &$\mathbf{\phantom{0}943.1\pm244.7}$ & $ \phantom{0}764.6\pm \phantom{0}96.6$ & $\phantom{0}624.0\pm175.0$ \\
& act. & $\phantom{0}609.0\pm167.6$ & $\mathbf{\phantom{0}994.1\pm 296.4}$ & $\phantom{0}550.6\pm124.0$ & $\mathbf{1060.9\pm299.3}$ & $ \phantom{0}890.8 \pm 108.7$ & $\phantom{0}724.7\pm181.7$ \\
& env. & $1241.6\pm438.7$ & $\mathbf{1971.6\pm545.3}$ & $1277.8\pm314.8$ & $\mathbf{2109.0\pm651.4}$ & $ 1856.8\pm 240.1$ & $1344.1\pm320.2$\\
& dom. & $ 1640.0\pm507.4$ & $\mathbf{1970.2 \pm 543.4}$ & $1277.1\pm317.2$ & $\mathbf{2111.3\pm650.8}$ & $ 1779.9\pm259.1$ & $1186.1\pm298.7$\\
\midrule
HalfCheetah & obs. & $1030.4\pm133.9$ & $ \mathbf{1325.4\pm\phantom{0}357.4}$ & $\phantom{0}956.4\pm\phantom{0}\phantom{0}59.2$ &$ 1116.6\pm\phantom{0}287.9$& $ \mathbf{1382.3\pm\phantom{0}579.7}$  & $\phantom{0}942.2\pm\phantom{0}87.0$ \\
& act. & $\phantom{0}773.1\pm 177.5$& $ \mathbf{1099.4\pm\phantom{0}399.3}$ & $\phantom{0}771.2\pm\phantom{0}160.6$ & $\mathbf{\phantom{0}986.2\pm\phantom{0}412.9}$ & $\mathbf{1316.9\pm\phantom{0}782.1}$  & $\phantom{0}685.5\pm166.6$\\
& env. & $1989.0\pm658.2$ & $\mathbf{3213.8 \pm 1417.0}$ & $\mathbf{2837.1 \pm \phantom{0}937.6}$ & $\mathbf{3014.3\pm 1109.5}$ & $\mathbf{2835.7\pm 1570.0}$ & $2202.6 \pm 927.7$\\
& dom. & $1871.0\pm583.5$ & $\mathbf{3035.4\pm 1309.8}$ & $ 2478.3\pm1129.5$ & $\mathbf{2730.3\pm1405.9}$ & $\mathbf{2677.5\pm 1454.6}$ & $2035.4\pm756.4$\\
\midrule
Pendulum & obs. & $129.5\pm\phantom{0}32.5$ & $\mathbf{146.4\pm 39.8}$ & $\mathbf{148.1\pm\phantom{0}74.9}$ & $\mathbf{177.4\pm46.2}$ &  $121.6\pm 22.9$ & $\phantom{0}57.4\pm\phantom{0}58.4$\\
& act. & $\mathbf{362.7\pm\phantom{0}43.7}$& $\mathbf{389.4\pm18.9}$ & $287.4\pm122.3$ & $302.5\pm80.1$ & $\mathbf{363.6\pm26.8}$ &  $156.8 \pm 165.1$ \\
& env. & $995.4\pm\phantom{00}8.0$ & $\mathbf{998.7\pm\phantom{0}4.2}$ &  $820.6\pm344.9$ & $975.8\pm24.7$ & $\mathbf{999.2\pm\phantom{0}3.6}$ & $381.8 \pm 401.4$\\
& dom. & $ 942.2\pm177.6$ &$\mathbf{998.7\pm\phantom{0}4.0}$ & $820.5\pm345.0$& $975.4\pm25.2$ & $\mathbf{999.9\pm\phantom{0}0.3}$ & $406.3\pm430.1$\\
\midrule
D-Pendulum & obs.& $\phantom{0}923.3\pm \phantom{00}61.5$ & $1014.1\pm\phantom{0}78.4$ & $\mathbf{1051.3\pm151.0}$ & $\mathbf{1093.8\pm121.6}$ & $ \phantom{0}956.1\pm\phantom{0}66.9$ & $\phantom{0}754.1\pm\phantom{0}\phantom{0}82.6$ \\
& act. & $2110.7\pm\phantom{00}74.6$ & $\mathbf{2160.2\pm\phantom{0}44.2}$  & $2016.9\pm\phantom{0}70.8$ & $1944.1\pm\phantom{0}55.9$ & $\mathbf{2168.5\pm\phantom{0}66.3}$ & $1917.0\pm\phantom{0}\phantom{0}64.6$ \\
& env. & $ 7494.9\pm1189.1$ & $\mathbf{8036.7\pm194.5}$ & $ \mathbf{7997.1\pm283.9}$ & $7856.8\pm601.9$ &  $5622.9\pm929.1$ & $3147.3\pm1157.1$ \\
& dom. & $7013.8\pm\phantom{0}850.6$ & $ \mathbf{8032.1\pm188.8}$ & $\mathbf{8006.9\pm278.6}$ & $7844.4\pm654.8$ & $  7436.7\pm696.6$ & $6667.0\pm1233.4$ \\
\bottomrule
\end{tabular}
\end{sc}
\end{small}
\end{center}
\vskip -0.1in
\end{sidewaystable*}

\end{document}